\documentclass{IOS-Book-Article}
\usepackage{mathptmx}
\usepackage{graphicx}
\usepackage{amsmath}
\usepackage{amssymb}
\usepackage{amsfonts}
\usepackage{amsthm}
\usepackage{url}
\usepackage{bbm}
\usepackage{makecell}
\usepackage{lipsum}

\def\hb{\hbox to 10.7 cm{}}

\newtheorem{prop}[]{Proposition}
\newtheorem{definition}[]{Definition}
\newtheorem{theorem}[]{Theorem}

\newcommand\blfootnote[1]{%
  \begingroup
  \renewcommand\thefootnote{}\footnote{#1}%
  \addtocounter{footnote}{-1}%
  \endgroup
}

\begin{document}

\def\thepage{}

 \begin{frontmatter}              

\title{Who wants accurate models? \\Arguing for a different metrics to take classification models seriously}

\author[A]{\fnms{Federico} \snm{Cabitza}\thanks{Both authors contributed equally to this work. Corresponding author: Federico Cabitza, federico.cabitza@unimib.it.}}
\author[A,B]{\fnms{Andrea} \snm{Campagner}}

\address[A]{University of Milano-Bicocca, Milano, Italy}
\address[B]{IRCCS Istituto Ortopedico Galeazzi, Milano, Italy}


     \begin{abstract}
     With the increasing availability of AI-based decision support, there is an increasing need for their certification by both AI manufacturers and notified bodies, as well as the pragmatic (real-world) validation of these systems. Therefore, there is the need for meaningful and informative ways to assess the performance of AI systems in clinical practice. Common metrics (like accuracy scores and areas under the ROC curve) have known problems and they do not take into account important information about the preferences of clinicians and the needs of their specialist practice, like the likelihood and impact of errors and the complexity of cases. In this paper, we present a new accuracy measure, the \emph{H-accuracy} ($Ha$), which we claim is more informative in the medical domain (and others of similar needs) for the elements it encompasses. We also provide proof that the H-accuracy is a generalization of the balanced accuracy and establish a relation between the \emph{H-accuracy} and the \emph{Net Benefit}. Finally, we illustrate an experimentation in two user studies to show the descriptive power of the $Ha$ score and how complementary and differently informative measures can be derived from its formulation (a Python script to compute $Ha$ is also made available). 
     \end{abstract}

\end{frontmatter}


\section{INTRODUCTION}
\label{sec:introduction}

It\blfootnote{Code to compute the Ha score available at \url{https://github.com/AndreaCampagner/uncertainpy}} has become a truism that the interest in medical AI has risen markedly in the last few years, with the accumulation of studies that, mostly in experimental settings and on retrospective data (e.g.~\cite{esteva2017dermatologist,gulshan2016development}), but also prospectively (e.g.~\cite{abramoff2018pivotal}), demonstrate that Machine Learning (ML) models, and the decision support systems integrating these models, can achieve a level of discriminative and predictive performance on a par with human clinicians, or even a slightly higher performance on very specific tasks~\cite{Liu2019,shen2019artificial}. 

A higher interest attracts more funding, and this in its turn allows for more market supply of systems that, nevertheless, to be marketed need to be validated and certified by competent bodies~\cite{cabitza2019proof}. This sheds light on, and attaches a renovated importance for, any valid and reliable measure of the performance of decision support applications that utilize machine-learning-based models, both to facilitate health technology assessment~\cite{goodman1999methodological}, product precertification and version-based approval renewals that deal with the continuous evolution of these systems, and also to give prospective users insights on the practical usefulness of these otherwise ``opaque'' and difficult-to-scrutinize systems.

Over time, the scholarly and professional community of statistical and machine learning has developed many metrics to assess and report the performance of discriminative models~\cite{sokolova2009systematic}: some metrics are common and easy-to-comprehend, like accuracy, recall (also known as sensitivity), specificity, the $F_{1}$ score (i.e., the harmonic mean of sensitivity and specificity), and the area under the ROC curve (AUROC); others are less common and less straightforward, like the Youden index (or Informedness), the Cohen’s Kappa, the Matthews correlation coefficient (i.e., the geometric mean of other Markedness and Informedness), the Gini coefficient (that is derived from the AUROC), the Kolmogorov-Smirnov measure and the logarithmic and Hamming loss. Also recent recommendations developed for the reporting of prediction models, like TRIPOD (Transparent reporting of a multivariable prediction model for individual prognosis or diagnosis~\cite{collins2015transparent}), do not generally recommend any specific measure, despite their comprehensiveness (the above TRIPOD encompasses a 22-item checklist).

In lack of more specific reporting guidelines, and notwithstanding the variety of metrics available (or right because of it), in the medical literature one measure of model performance is much more common than the others, and this is \emph{accuracy}: the proportion of correct predictions with respect to the total number of predictions made. Accuracy is obviously related to error rate (that is the complementary concept), and represents the capability of the model to provide the decision maker with the right recommendation for any case at hand. 

``Getting the right answer'' seems ``a good standard to use to judge the quality of a decision support system [but] there are problems with this criterion'' as said in~\cite{berner2003diagnostic}. Indeed, despite its apparent simplicity, accuracy scores can be misleading or, at least, do not tell the full story. For instance, it is generally well known that the accuracy of an ML model must be evaluated on a test set, that is, on different instances from those used to train the model, otherwise its high values could mirror over-fitting rather than informing about the actual goodness of the model. 

\begin{figure*}[bth]
    \centering
    \includegraphics[width=\textwidth]{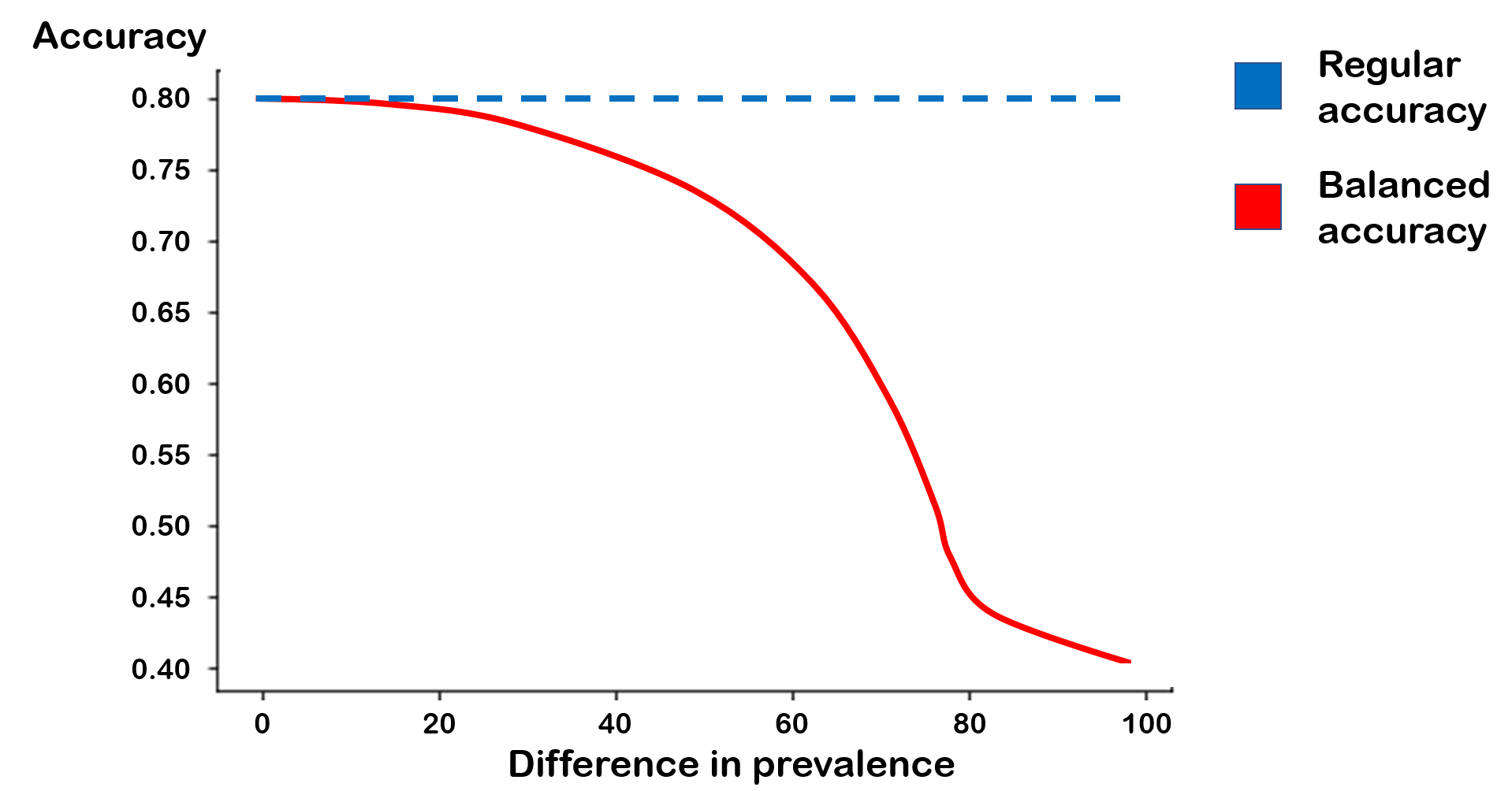}
    \caption{Regular accuracy, here conventionally set at 80\%, and balanced accuracy coincide if the dataset is perfectly balanced; however, as class prevalences differ, so different measures of accuracy do. Mind that at a certain point the difference in prevalence is combined with a difference in the accuracy for each class.}
    \label{fig:balanced}
\end{figure*}

Another potential source of bias lies in highly imbalanced test datasets, for which a so-called \emph{accuracy paradox} has been observed, that is when ``high accuracy is not necessarily an indicator of high classifier performance'' \cite{albacete2014accuracy}. Thus, in case of datasets where classes are unevenly represented, alternative measures have been proposed instead of regular accuracy, like the \emph{Matthew correlation coefficient}~\cite{chicco2017ten} and the \emph{balanced accuracy}~\cite{brodersen2010balanced}: in particular, this latter is intuitively defined as the average of the accuracy obtained on each class individually. To illustrate this point, in Figure~\ref{fig:balanced} we show how increasing the difference in the prevalence of the two classes of a binary discriminative task makes regular and balanced accuracy diverge progressively, so that the meaningfulness of the former and more common type of accuracy, which is blind to class prevalence problems, can be disputed\footnote{The same problem affects the area under the ROC curve, another very common performance metrics in medical AI that is criticized also for other problems, like the one of averaging the performance of a model on (operating) points that lack real utility or real-world significance~\cite{pencina2015evaluating}.}~\cite{albacete2014accuracy}.
Moreover, not only the regular accuracy, but also the balanced accuracy neglect other meaningful aspects of the latent variation of the training datasets, a problem that has been recently denoted as \emph{hidden stratification}~\cite{oakden2019hidden}: accuracy can be high for a ML model, but this latter still misses \emph{relevant} findings, that is conditions that can have a strong impact on the patient's outcome.

We also believe that scores of regular accuracy (and partly also those of balanced accuracy) are only a partial way to convey meaningful information on the usefulness and practical value of a predictive model: this is mainly because these metrics overestimate the model performance and do not take into account important characteristics of the test set (besides class imbalance): the importance of rightly detecting one class at the expense of the other(s), according to the users' preferences; the extent the model gets the right answer by chance (and hence its intrinsic reliability); and the \emph{complexity} of the cases correctly identified (and hence the difficulty of the discrimination for a human rater).

For this reason, in this paper we present a new accuracy measure, and related metrics, the H-accuracy, or $Ha$, which we claim is more informative in the medical domain (and other domain with similar requirements) as it covers all of the aspects mentioned above. We show the descriptive power of our proposal on a well-known dataset used to compare and evaluate medical artificial intelligence in the radiological domain, and also provide proof that the H-accuracy is a generalization of balanced accuracy and hence a proper measure of accuracy alternative to regular accuracy, and whose properties make it suitable to assess the ``skills'' of predictive ML models.

\section{METHOD}
\label{sec:methods}

In what follows, we will describe the method by which to derive a multi-parametric (and hence multi-dimensional) type of classification accuracy that we propose to better mirror the rich and multi-faceted meanings the original construct of accuracy has in \emph{naturalistic decision making}~\cite{klein2008naturalistic}; and the method by which we designed the user studies through which we derived the parameters empirically by involving experts and prospective users of the decision support to be evaluated in terms of accuracy.

\subsection{The H-accuracy metrics}

In what follows we report the formula for the H-accuracy (see Equation~\ref{eq:ha}). Assume for the following that we deal with a classification task in which the set of classes (or labels) is $C = \lbrace l_1, ..., l_k \rbrace$; we have a dataset of instances $S = \lbrace x_1, ..., x_n \rbrace$ sampled from the space of all possible osservations $X$ and the real class of instance $x$ is given by a function $c : X \mapsto C$. We consider as possible predictive models only \emph{scoring models} (notice that this assumption is not restrictive, given that most Machine Learning (ML) models are, or can be defined as, scoring models), that is a model $m$ is defined as $m: X \mapsto [0,1]^k$ where, given $m(x) = \langle l_{1_{m}}(x), ..., l_{k_{m}}(x) \rangle$, $l_{i_{m}}(x)$ represents the confidence that model $m$ attaches to the event $c(x) = l_i$, thus we also define $\mathbf{m}(x) = argmax_{l_{i}} l_{i_{m}}(x)$ (i.e. the class in $C$ for which $m$ has maximum confidence, which is traditionally what we assume $m$ provides if asked for explicitly labeling an instance $x$).
Under this framework, \emph{accuracy} (what we will term \emph{regular} accuracy) of a model $m$ is defined as:
\begin{equation}
    a(m) = \frac{1}{|S|}\sum_{x \in S} \mathbbm{1}_{\mathbf{m}(x) = c(x)}
\end{equation}

where $\mathbbm{1}$ is the indicator function.
\emph{Balanced accuracy}, on the other hand, is defined as:
\begin{equation}
    Ba(m) = \frac{1}{|C|}\sum_{l \in C}\frac{1}{|\lbrace x \in S : c(x) = l \rbrace|}\sum_{x \in S : c(x) = l} \mathbbm{1}_{m(x)=c(x)}
\end{equation}

We then introduce the formula for the H-accuracy of a model $m$ as:

 \begin{equation}
 \begin{split}
      Ha(m | \tau, \mathbf{p}, \mathbf{d})  =  \sum_{l \in C} \left(\mathbf{p}(l)*
            \sum_{x \in S : c(x) = l} \frac{\mathbf{d}(x)}{\sum_{c(x) = l}\mathbf{d}(x)}*\sigma_m(\mathbf{l}(x) | \tau) \right)
      \label{eq:ha}
      \end{split}
  \end{equation}
  
 \noindent
  where: 
  \begin{enumerate}
      \item $\mathbf{p}(l)$ is the weight (or \emph{priority}) associated to label $l \in C$ (with $\sum_{l \in C}\mathbf{w}(l) = 1$), that is a representation of the relative importance attached to the fact that the model correctly predicts that label; 
      \item $\mathbf{d}(x) \in [0,1]$ is the complexity of instance $x$, that is the intrinsic (i.e., independent of the single decision maker) difficulty of making the correct prediction about the case $x$;
      \item $\mathbf{l}(x)$ is the prediction score for class $l$ on instance $x$;
      \item $\sigma(\mathbf{l}(x) | \tau)$ is a penalty function defined as\footnote{Notice that when $\tau = \frac{1}{|C|}$ the function is defined as $\sigma(\mathbf{l}(x)|\tau) = \begin{cases}
  0 & \mathbf{l}(x) < max_{c \in C} \mathbf{c}(x)\\
  1 & \mathbf{l}(x) \geq max_{c \in C} \mathbf{c}(x)
  \end{cases}$}:
  \begin{equation}
  \label{eq:sigma}
  \sigma(\mathbf{l}(x) | \tau) = 
  \begin{cases}
  0 & \mathbf{l}(x) < max_{c \in C} \mathbf{c}(x)\\
  \frac{\mathbf{l}(x) - \frac{1}{|C|}}{\tau - \frac{1}{|C|}} & max_{c \in C} \mathbf{c}(x) \leq \mathbf{l}(x) \leq \tau \\
  1 & \mathbf{l}(x) > \tau
  \end{cases}
  \end{equation}
  
    \end{enumerate}
  
  Before analyzing the terms in $Ha$ in more details, we first prove that $Ha$ is a proper generalization of the balanced accuracy (Ba) and, thus, a proper accuracy measure:
  \begin{prop}
Let $C = \lbrace l_1, ..., l_k \rbrace$ $\tau = \frac{1}{|C|}$, $\mathbf{p}(l_1) = ... = \mathbf{p}(l_k) = \frac{1}{|C|}$ and $\forall x \in X. \mathbf{d}(x) = k$. Then $\forall h . Ba(m) = Ha(m | \tau, \mathbf{p}, \mathbf{d})$.
  \label{pro:balanced}
  \end{prop}
  \begin{proof}
  \begin{equation}
  \begin{split}
      Ha(m | 0.5, \frac{1}{|C|}, k) = \frac{1}{|C|}\sum_{l \in C}\sum_{x \in S : c(x) = l} \frac{k}{k*|\lbrace x \in S : c(x) = l \rbrace|}\mathbbm{1}_{m(x) = l}
      \end{split}
\end{equation}

from which, by rearranging terms and simplifying, we obtain the statement.

\end{proof}

In what follows, we explain each term in the definition of the $Ha$ measure in greater detail:
\begin{enumerate}
    \item Consider first term $\sigma(\mathbf{l}_m(x)|\tau)$; this terms represents a penalty which is assigned to the model $m$ for its prediction on instance $x$ whose real label is $l$, by assigning a lower relevance in the accuracy computation to this instance if the confidence of the model in its prediction is too small (i.e. lower than $\tau$, which represents a confidence threshold). First of all notice that for regular accuracy (and also balanced accuracy) we have that $\tau = \frac{1}{|C|}$ $\sigma(\mathbf{l}_m(x) | \frac{1}{|C|}) = 1$ if $\mathbf{l}_m(x) = max_{c \in C} \mathbf{c}(x)$ and 0 otherwise, thus, to be meaningful, it must always hold that $\tau \geq \frac{1}{|C|}$. Furthermore, according to the definition of the $\sigma$ function as in Equation \ref{eq:sigma}, when the model does not have enough confidence in its best prediction (i.e. $\mathbf{l}_m(x) < \tau$) the penalty function provides a linear interpolation between 0 and 1 on the prediction score given to class $l$ for instance $x$;
    \item The $\sigma$ component of the formula, described in the previous point, is then reweighed by the term $\frac{\mathbf{d}(x)}{\sum_{c(x) = l}\mathbf{d}(x)}$, which essentially weighs an instance $x$ by its relative difficulty (normalized) thus assigning greater importance on more difficult instances in the computation of the accuracy;
    \item The partial accuracy due to a specific class $l \in C$ (i.e. the inner sum in Equation \ref{eq:ha}) is then weighed by $\mathbf{p}(l)$ which represents the importance (e.g. defined by the prospective users of the predictive model) of correctly predicting on instances whose real class is $l$ (and thus a preference towards avoiding mis-classifications for $l$ instances) possibly sacrificing some accuracy on other classes. Notice, for example, that in balanced accuracy $\forall l \in C . \mathbf{p}(l) = \frac{1}{|C|}$ because we have, by default, no preference in favor of any specific classes. Thus, the effect of the external summation is to compute a weighed average of the $l$-specific contributions to the accuracy in which the weighing factors are defined by $\mathbf{p}$.
\end{enumerate}

\begin{figure*}[!thb]
    \centering
    \includegraphics[width=\textwidth]{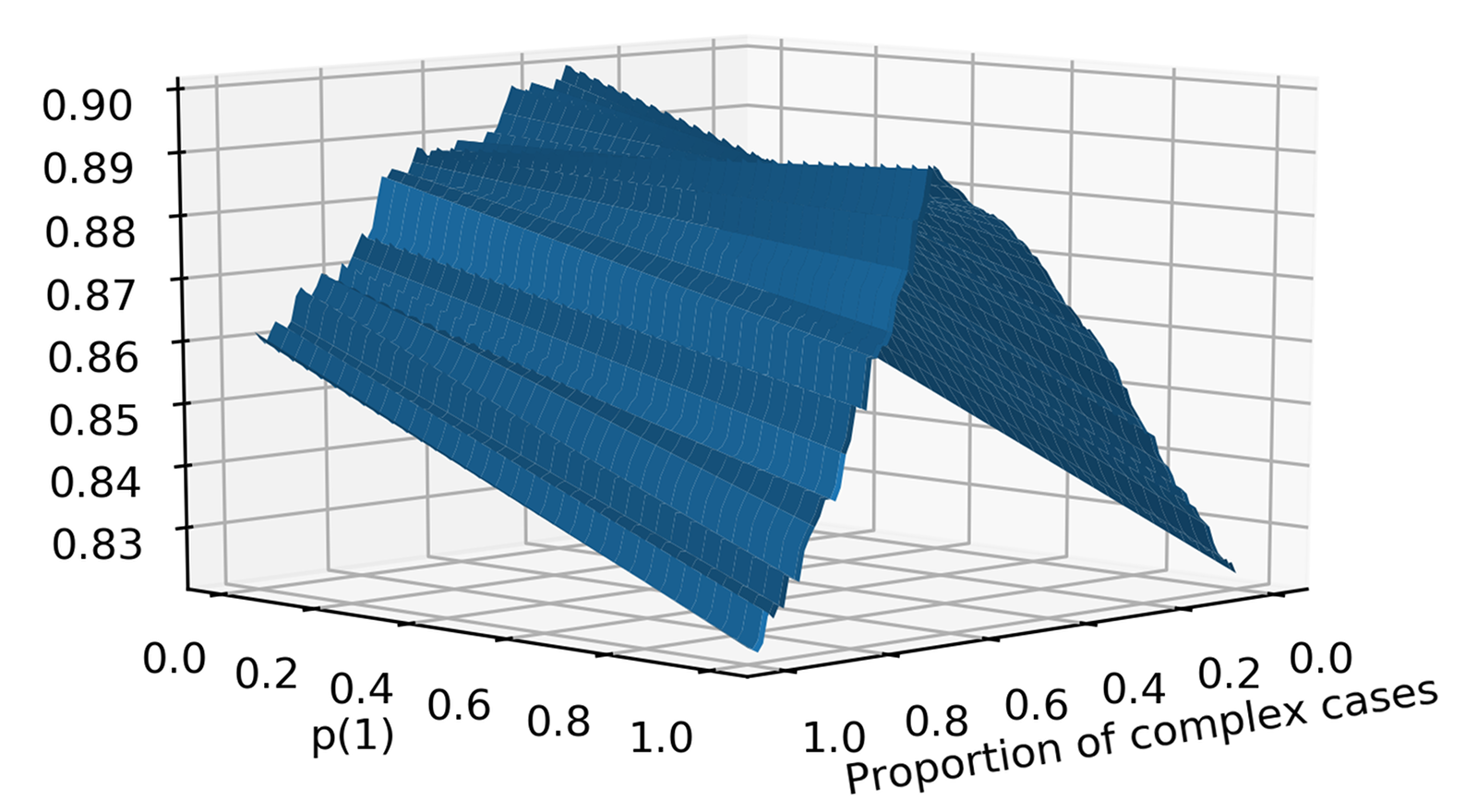}
    \caption{The surface profile of the $Ha$ function with respect to the proportion of complex cases and the priority assigned to the positive class (1, in our case, abnormal imaging with some lesion). See also Figure~\ref{fig:complex} to see the projection of the surface on the accuracy/complexity plane.}
    \label{fig:ha}
\end{figure*}

A 3D visualization of the $Ha$ function, for binary classification tasks, with respect to the priority of the positive class $\mathbf{p}(1)$ and the proportion $d$ of complex cases (this proportion, is in turn used to randomly sample a set of specific instances of the $\mathbf{d}$ function, which are then average to obtain a smoother profile surface) is shown in Figure \ref{fig:ha}.

If, for the sake of the argument, we assume the equivalence in semantics of $\tau$ in both formulations, we can relate the H-accuracy also with the \emph{Net Benefit}~\cite{vickers2006decisioncurve}. This latter one is defined as:
\begin{equation}
    \label{eq:netbenefit}
    NB(\tau) = TPR_\tau*\pi - (1 - \pi)*\frac{\tau}{1 - \tau}FPR_\tau
\end{equation}

where $\tau$ is a probability threshold (sometimes called risk), $\pi$ is the prevalence of the positive class, $TPR_\tau$ and $FPR_\tau$ are the values of the true positive rate and false positive rate at a fixed probability threshold $\tau$.

\begin{theorem}
\label{th:nbha}
Let \begin{equation}
\label{eq:sigma_alt}
\sigma_{risk}(\mathbf{l}(x)|\tau) = 
\begin{cases}
1 & (\mathbf{l}(x) \geq \tau \wedge \mathbf{l} = 1) \vee \mathbf{l}(x) > (1 - \tau) \wedge \mathbf{l} = 0\\
0 & otherwise
\end{cases}
\end{equation}.

Then $NB(\tau) = \frac{Ha(\tau, \mathbf{p}=\langle \frac{\tau*(1 - \pi)}{\alpha}, \frac{(1 - \tau)*\pi}{\alpha} \rangle, \mathbf{d}=k), \sigma_{risk}}{1 - \tau}*\alpha - \frac{\tau*(1 - \pi)}{1 - \tau}$, where $\alpha$ is a  normalization factor.
\end{theorem}
\begin{proof}
$NB(\tau) = TPR_\tau*\pi - (1 - \pi)*\frac{\tau}{1 - \tau}FPR_\tau = TPR_\tau*\pi - (1 - \pi)*\frac{\tau}{1 - \tau}(1 - TNR_\tau)$. Then 
\begin{equation}
(1 - \tau)*NB(\tau)= (1- \tau)*\pi*TPR_\tau + \tau*(1 - \pi)*TNR_\tau - \tau*(1 - \pi)
\end{equation}

If we let $\alpha = \tau*(1 - \pi) + (1 - \tau)*\pi $ be a renormalization factor, then

\begin{equation}
 \\ \frac{(1 - \tau)*NB(\tau)}{\alpha}= Ha(\tau, \mathbf{p}: \langle \frac{\tau*(1 - \pi)}{\alpha}, \frac{(1 - \tau)*\pi}{\alpha} \rangle, \mathbf{d}: k, \sigma_{risk}) - \frac{\tau*(1 - \pi)}{\alpha}
 \end{equation}
hence the result.
\end{proof}

In particular, setting $\pi = 0.5$, $$NB(\tau=0.5) = Ha(0.5, \mathbf{p} = 0.5, \mathbf{d} = k) - 0.5 =  Ba - 0.5 = \frac{J(\tau=0.5)}{2}$$ where $J$ is the \emph{Youden index}~\cite{youden1950youdenindex}. Using the same technique, a similar relationship can be established between the $Ha$ and the \emph{standardized Net Benefit} $sNB = \frac{NB}{\pi}$ (also known as \emph{relative utility})~\cite{baker2009riskprediction,baker2009relativeutility,baker2015evalru}: in particular when $\pi = 0.5$ it holds $$(1 - \tau)*sNB(\tau) = Ha(\tau, \mathbf{p}: \langle \tau, 1 - \tau \rangle, \mathbf{d} : k, \sigma_{risk}) - \tau$$.
We will see in Section \ref{sec:discussion} why the relation between the $Ha$ and the $NB$ is of interest for the comprehensive evaluation of decision support systems. 

\subsubsection{Relevant instances of Ha}

In what follows we define three relevant instances of $Ha$ by focusing on each of its parameters $\mathbf{p}(l)$, $\mathbf{d}(x)$ and $\mathbf{\tau}$, and fixing the other variables, to some constant. 

Thus we define \emph{prioritized accuracy} at a specific $p$, by means of Formula~\ref{eq:ha} ($Ha(\mathbf{p}) := Ha(m|\frac{1}{|C|}, \mathbf{p}, k)$), that is an accuracy that takes into account the perceived differential importance of avoiding different kinds of error by the decision support in diagnostic practice. When a panel of raters are also called to re-classify an already established ground truth, preferences can be set considering their classification performance. For instance, looking at a ROC space, as the one depicted in Figure~\ref{fig:ROC} one can notice whether the points are placed closer to the left side of the chart than to the right side or not, thus indicating a preference for  specificity or sensitivity, respectively. Analytically, the value of $p$ can be determined as described in Formula~\ref{eq:priority}, by computing the average \emph{true positive rates} and \emph{true negative rates} (which, in ROC space, are respectively the y coordinate and the opposite of the x coordinate, i.e. $1 - x$) and then normalizing in order to satisfy the constraint that the priorities must sum to 1.

\begin{equation}
\label{eq:priority}
\begin{split}
    p(0) &= \frac{mean(TNR)}{mean(TPR) + mean(TNR)}\\
    p(1) &= \frac{mean(TPR)}{mean(TPR) + mean(TNR)}
\end{split}
\end{equation}

Complexity allows us to define \emph{practical accuracy}: this is a score that takes into account the capability of the predictive model to provide the correct recommendation for the complex cases, that is when the decision maker needs its support more~\cite{meyer2013physicians}, assuming that in the easiest cases the decision makers could override the model's recommendation more easily and be less affected by \emph{automation bias}~\cite{bahner2008misuse,goddard2011automation} (that is the tendency of users to be misled by the wrong advice given by decision supports that they consider reliable). 

In particular in what follows, we will consider the distribution $\mathbf{d}$ of complexities obtained by considering the average complexity observed in a user study for each case and then transforming these averages according to a two-level complexity scale, by mapping cases with low to average complexity to a complexity value of $\frac{1}{2}$ and cases with high reported complexity to a value of $1$ (after this transformation, the complex cases, i.e. those with $d(x) = 1$ were approximately a third of the whole dataset).
In Section~\ref{sec:discussion}, we will see other possible ways to define $d(x)$.

 Lastly, also a \emph{confident accuracy} can be defined (i.e. $Ha(\tau) = Ha(m | \tau, \frac{1}{|C|}, k)$), that is an accuracy estimate that takes into account the error rate when the decision support is confident ``enough'' of its recommendation and that considers that, for all of the other cases (equated to cases in which the model ``takes a guess''), errors are just errors and right answers are penalized, proportionally with respect to the related confidence (or prediction score). To this aim, let us consider $\tau$ in Formula~\ref{eq:ha}: this can be seen as a sort of \emph{confidence threshold}, under which the model should abstain from giving advice and, if it does not abstain (as in traditional machine learning), its accuracy score gets penalized, so to weight out the role of chance that ``inflated'' it. 
 To set $\tau$ parameter in Formula~\ref{eq:ha}, we will normalize the (ordinal) confidence level observed in a user study and find, for each fraction $r \in [0, 1]$ of correctly classified cases, the maximum confidence level for which at least $r$ correctly classified cases corresponds (more details will be given in Section~\ref{sec:discussion}, where also alternative ways to set $\tau$ will be discussed).

\subsubsection{Relevant properties of Ha}

Lastly, in what follows we will consider the properties of the $Ha$ measure with respect to the \emph{invariance} laws defined in \cite{sokolova2009systematic}, which analyse different kinds of invariance of any accuracy measure with respect to transformations of the \emph{confusion matrix} (CM). We will focus on the binary case (i.e. $C = \lbrace 0, 1 \rbrace$, for which the CM can be expressed as:
\begin{equation}
    \begin{bmatrix}
    tp & fn\\
    fp & tn
    \end{bmatrix}
\end{equation}

where tp, fn, fp and tn represent, respectively, the \emph{true positives}, \emph{false negatives}, \emph{false positives} and \emph{true negatives}.

First of all, we can observe that in order to consider if the $Ha$ satisfies the mentioned invariance properties it is necessary to constrain the measure so that it is properly defined on a confusion matrix: this requires fixing $\tau = 0.5$, indeed if this is not the case then $Ha$ does not strictly operate on the confusion matrix (indeed, it can be seen that the basic equality $tp + tn + fp + fn = N$ may not hold) so in the following we will assume that $\tau = 0.5$ without explicitly pointing this out in all propositions. Furthermore, being $Ha$ a proper generalization of $Ba$ (and, also, of \emph{sensitivity} and \emph{specificity}, see \cite{sokolova2009systematic}) in its most general parametric form the measure needs only satisfy the following property:
\begin{definition}
An accuracy measure $f$ is $I_6$ invariant if it is invariant under uniform multiplicative changes, that is $f(\begin{bmatrix}
    tp & fn\\
    fp & tn
    \end{bmatrix}) = f(k\begin{bmatrix}
    tn & fp\\
    fn & tp
    \end{bmatrix})$
\end{definition}

\begin{prop}
Let $\mathbf{d} = k$ then $Ha(\mathbf{p})$ is $I_6$ invariant. 
\end{prop}
\begin{proof}
$Ha(\mathbf{p}) = \mathbf{p}(0)\frac{tn}{tn + fp} + \mathbf{p}(1)\frac{tp}{tp + fn} = \mathbf{p}(0)\frac{k*tn}{k*tn + k*fp} + \mathbf{p}(1)\frac{k*tp}{k*tp + k*fn}$
\end{proof}

However, for each invariance property $I_i$ (except $I_7$, as we will show below) it is possible, by appropriately setting $\mathbf{p}$,to obtain specific instances of the $Ha$ satisying at least property $I_i$.

\begin{definition}
An accuracy measure $f$ is $I_1$ invariant if it is invariant with respect to exchange of positive and negative classes, that is if $f(\begin{bmatrix}
    tp & fn\\
    fp & tn
    \end{bmatrix}) = f(\begin{bmatrix}
    tn & fp\\
    fn & tp
    \end{bmatrix})$
\end{definition}

\begin{theorem}
Let $\tau = 0.5$ and $\mathbf{d}(x) = k$. Then if $\mathbf{p}(1) = \frac{tp + fn}{N}$ and $\mathbf{p}(0) = \frac{fp + tn}{N}$ $Ha(\mathbf{p})$ is $I_1$ invariant by swapping $\mathbf{p}(0) \mathbf{p}(1)$ when exchanging classes.
\end{theorem}
\begin{proof}
In this case we have that $Ha(\mathbf{p}) = \mathbf{p}(0)\frac{tn}{tn + fp} + \mathbf{p}(1)\frac{tp}{tp + fn}$. Then we have $Ha(\mathbf{p}) = \frac{tn + tp}{N} = a$ which is $I_1$ invariant.
\end{proof}

\begin{definition}
An accuracy measure $f$ is $I_2$ invariant if it is invariant with respect to changes of $tn$; is $I_3$ invariant if it is invariant with respect to changes of $tp$; is $I_4$ invariant if it is invariant with respect to changes of $fn$; is $I_5$ invariant if it is invariant with respect to changes of $fp$; is $I_8$ invariant if it is invariant under rows' changes, that is $f(\begin{bmatrix}
    tp & fn\\
    fp & tn
    \end{bmatrix}) = f(\begin{bmatrix}
    k_1*tn & k_1*fp\\
    k_2*fn & k_2*tp
    \end{bmatrix})$
with $k_1 \neq k_2$.
\end{definition}

\begin{theorem}
Let $\tau = 0.5$ and $\mathbf{d}(x) = k$. Then if $\mathbf{p}(0) = 0$ $Ha$ is $I_2$ and $I_5$ invariant, if $\mathbf{p}(1) = 0$ $Ha$ is $I_3$ and $I_4$ invariant; in both cases it is $I_8$ invariant.
\end{theorem}
\begin{proof}
In the first case we have that $Ha = p(1)\frac{tp}{tp + fn}$, in the second case we have that $Ha = p(0)\frac{tn}{tn + fp}$ which, respectively, satisfy the two properties. Both instances of $Ha$ satisfy the $I_8$ invariance
\end{proof}

\begin{definition}
An accuracy measure $f$ is $I_2$ invariant if it is invariant under columns' changes, that is $f(\begin{bmatrix}
    tp & fn\\
    fp & tn
    \end{bmatrix}) = f(\begin{bmatrix}
    k_1*tn & k_2*fp\\
    k_1*fn & k_2*tp
    \end{bmatrix})$
with $k_1 \neq k_2$.
\end{definition}
Then, evidently: 
\begin{prop}
$Ha$ is not $I_7$ invariant.
\end{prop}

Notice, relevantly, that in \cite{sokolova2009systematic} any invariance property (and the negation of invariance $I_7$) is naturally associated with a specific Machine Learning setting for which satisfaction of the invariance property corresponds to soundness of an accuracy metrics. 
Notice, also, that while in the previous results we focused only on the case of constant complexity across the cases, similar results can be obtained under any complexity distribution $\mathbf{d}$, although in the general case it is more difficult to provide a simple, analytical expression for the appropriate values to give to the parameter $\mathbf{p}$.

\begin{figure*}[!bht]
    \centering
    \includegraphics[width=.8\textwidth]{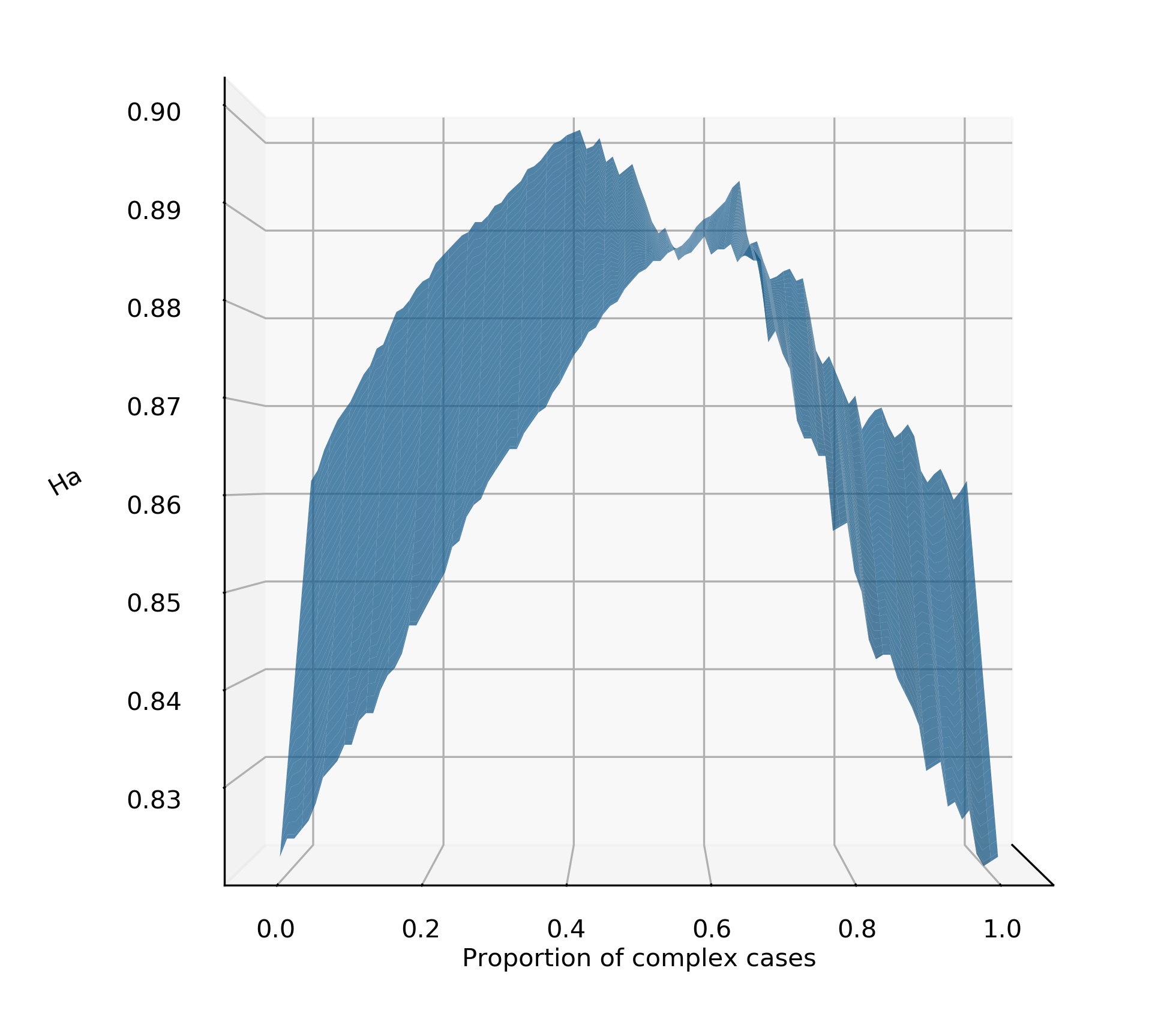}
    \caption{The behavior of the model performance (Ha) as a function of the proportion of complex cases. The accuracy degradation both by increasing and decreasing the number of complex cases can be explained with the training being optimized for a specific proportion of complex cases, and for other potential hidden stratifications.}
    \label{fig:complex}
\end{figure*}

\subsection{The empirical setting}

In this section we report about two user studies. In the first study we involved 31 doctors working in one of the largest Italian teaching hospitals specialized in musculoskeletal disorders, the IRCCS Orthopedics Institute Galeazzi of Milan. In the second study, we involved 13 radiologists from the same hospital, by asking them to annotate a sample of 417 cases randomly extracted from the MRNet dataset\footnote{\url{https://stanfordmlgroup.github.io/competitions/mrnet/}.}, which encompasses 1,370 knee MRI exams performed at the Stanford University Medical Center (with 81\% abnormal exams, and in particular 319 Anterior Cruciate Ligament (ACL) tears and 508 meniscal tears). Our sample was balanced with respect to abnormal cases and type of abnormality.

In both studies we used an online questionnaire platform (Limesurvey, version 3.18~\footnote{https://www.limesurvey.org/}) and invited the participants by personal email. In the first study, we asked the doctors to indicate: 1) what the minimum acceptable accuracy an ML model should exhibit in order to be clinically useful in their diagnostic decision making; 2) whether they preferred to be supported by a predictive model that was configured to optimize either sensitivity, or specificity, or also that was less accurate but ``privileged'' neither sensitivity or specificity. While the questionnaire administered in this first study was longer and encompassed further items that we will not consider in this paper, we will focus on these two items, by which we collected the necessary preferences to set the parameters of $Ha$. 

The second study was more articulated. As anticipated above, we involved 13 radiologists (of different MRI reading skills, which we stratify in two subgroups, higher- and lower-proficiency according to self-assessment) in a diagnostic task where they were called to discriminate the MRNet cases that were positive, and indicate whether these regarded either ACL or meniscal tears: in particular they had to say whether the presented imaging presented a case of ACL tear (yes/no), or a meniscal tear (yes/no): hence two decisions in total.
The radiologists were also requested to assess each case in terms of complexity (or difficulty in giving the correct answer) on an ordinal scale, and the confidence with which they classified the case, on an ordinal scale as well.

Also this study was conceived to get indications for the parametrization of the $Ha$ construct in Formula~\ref{eq:ha}.

\section{RESULTS}

    

\begin{table*}[tbh]
    \centering
    \caption{The accuracy of each rater involved in the empirical study in detecting ACL tears, meniscal tears, and for all of the cases.}
    \setlength\tabcolsep{4pt}
    \begin{tabular}{|c|c|c|c|c|}
    \hline
          & Min & Max & Avg & Std\\\hline
         Acc. ACL & 0.83 & 0.88 & 0.86 & 0.02\\\hline
         Acc. Men & 0.78 & 0.83 & 0.80 & 0.02 \\\hline
         Acc. Tot & 0.78 & 0.86 & 0.82 & 0.02\\\hline
    \end{tabular}
    \label{tab:acc}
\end{table*}



\begin{figure*}[bht]
    \centering
    \includegraphics[width=\textwidth]{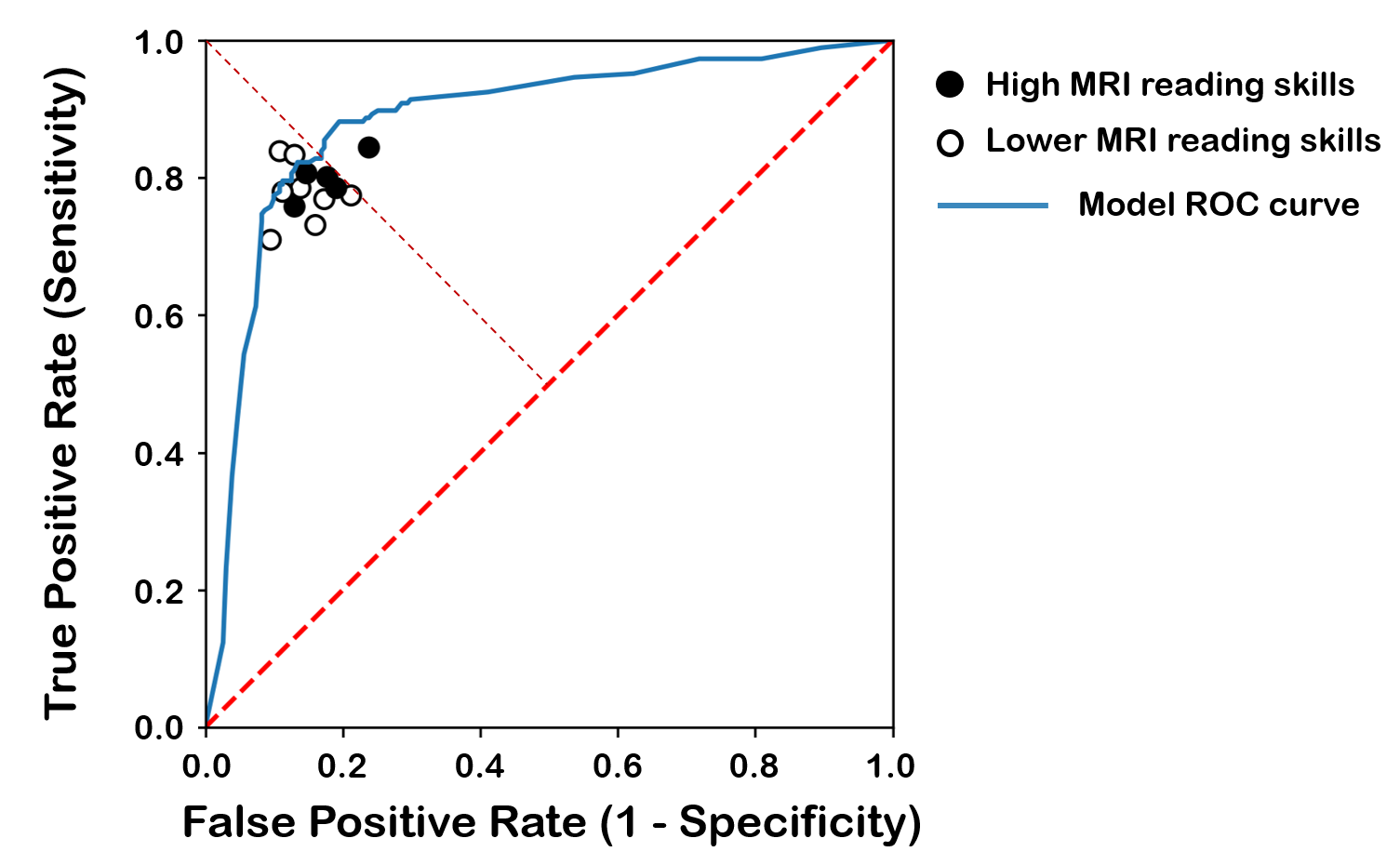}
    \caption{The radiologists diagnostic performance on the MRNet dataset in terms of TPR and FPR; full circles indicate expert radiologists; empty circles lower-expertise MRI readers (self-assessed). The blue line represents the ROC curve of our predictive model.}
    \label{fig:ROC}
\end{figure*}

\begin{figure*}[bth]
    \centering
    \includegraphics[width=\textwidth]{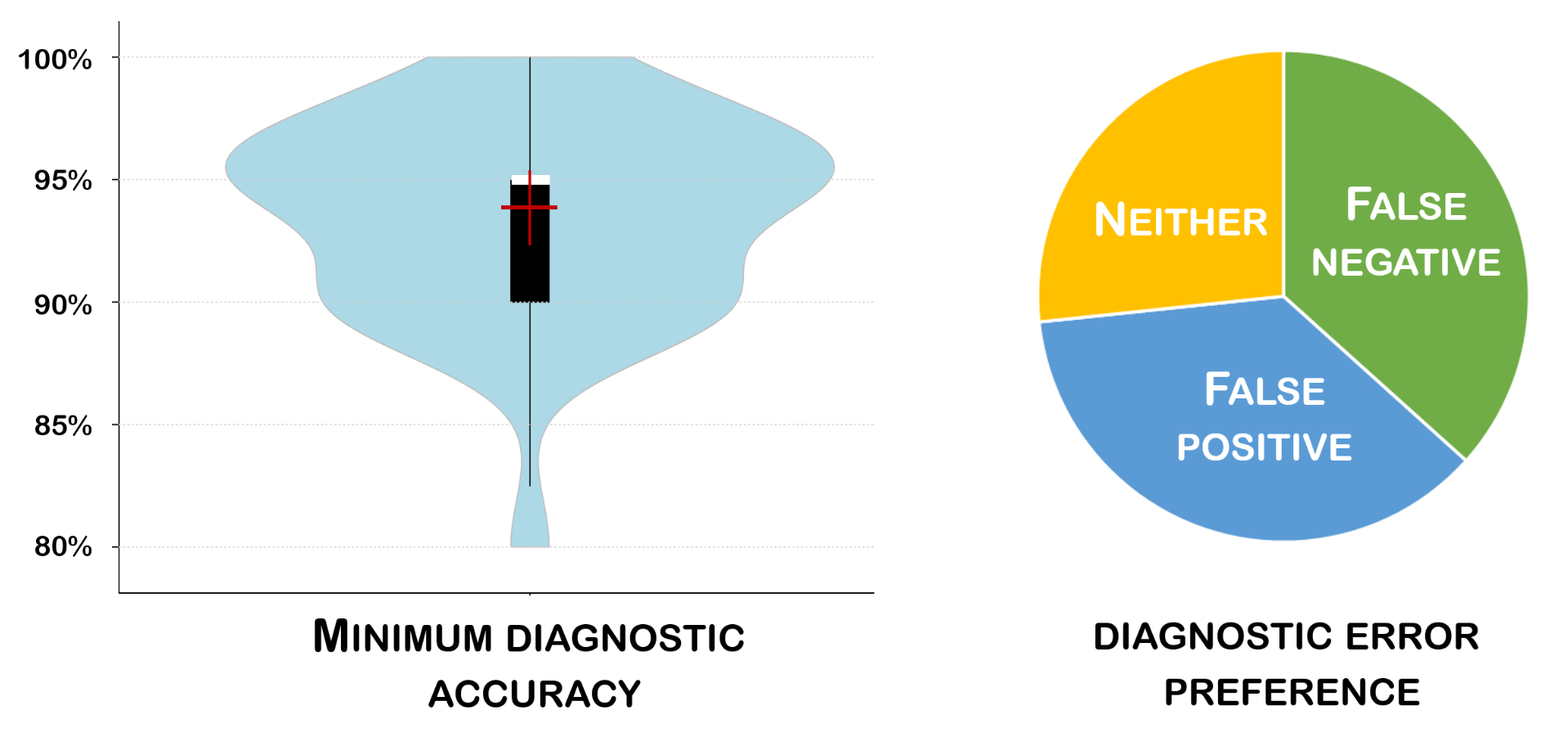}
    \caption{From left to right, the minimum diagnostic accuracy (the mean accuracy is indicated in red with its 95\% confidence interval); and the preferences about what kind of accuracy is more important at class level. In both cases, N=31.}
    \label{fig:res}
\end{figure*}

\begin{figure*}[tbh]
    \centering
    \includegraphics[width=\textwidth]{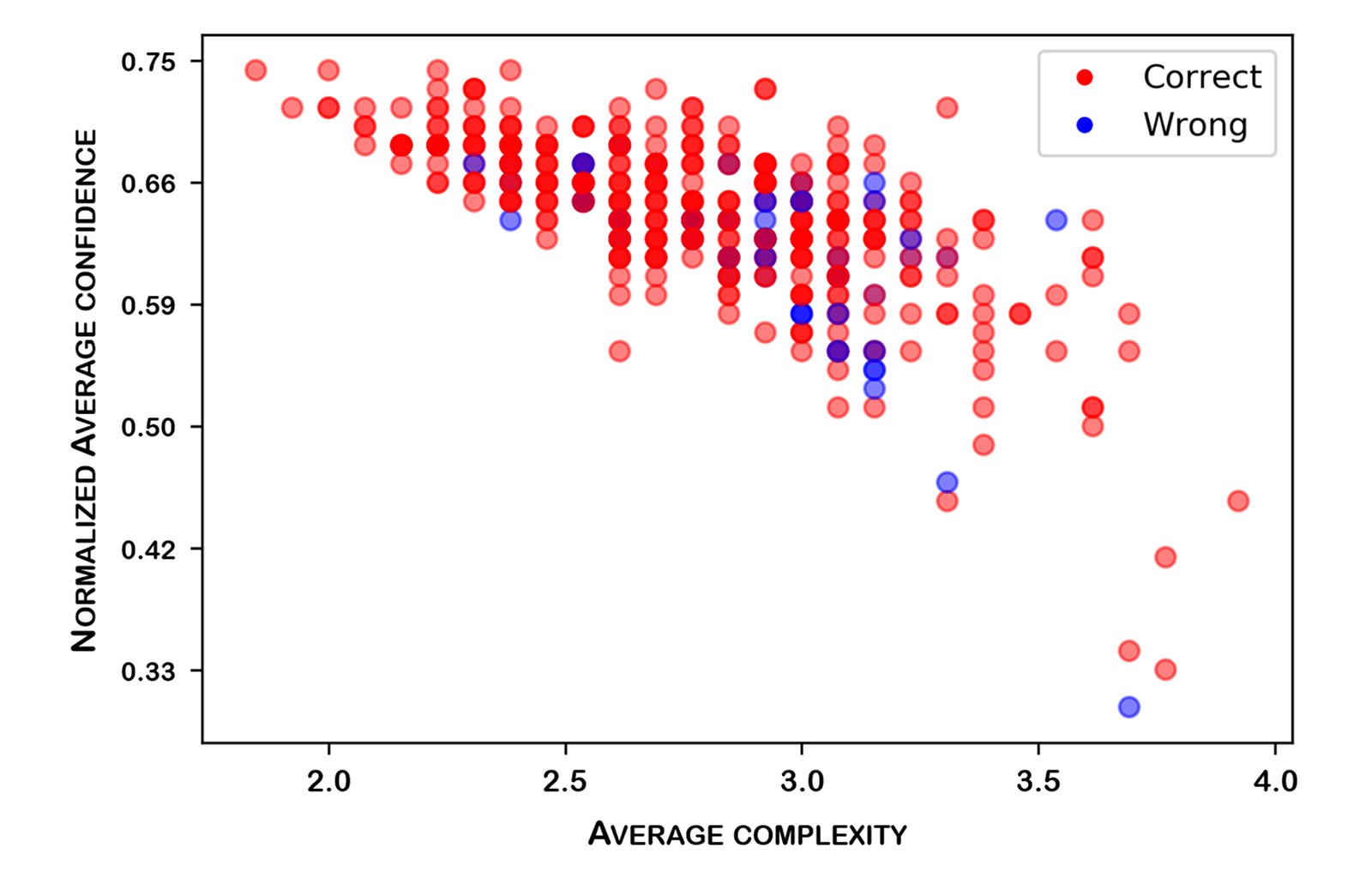}
    \caption{A scatter plot showing the negative correlation between the average normalized confidence of the ratings for each case and the average perceived complexity of the cases labelled. We also report whether the cases were rated correctly or not (in red and blue, respectively).}
    \label{fig:scatter}
\end{figure*}

\begin{figure*}[bth]
    \centering
    \includegraphics[width=\textwidth]{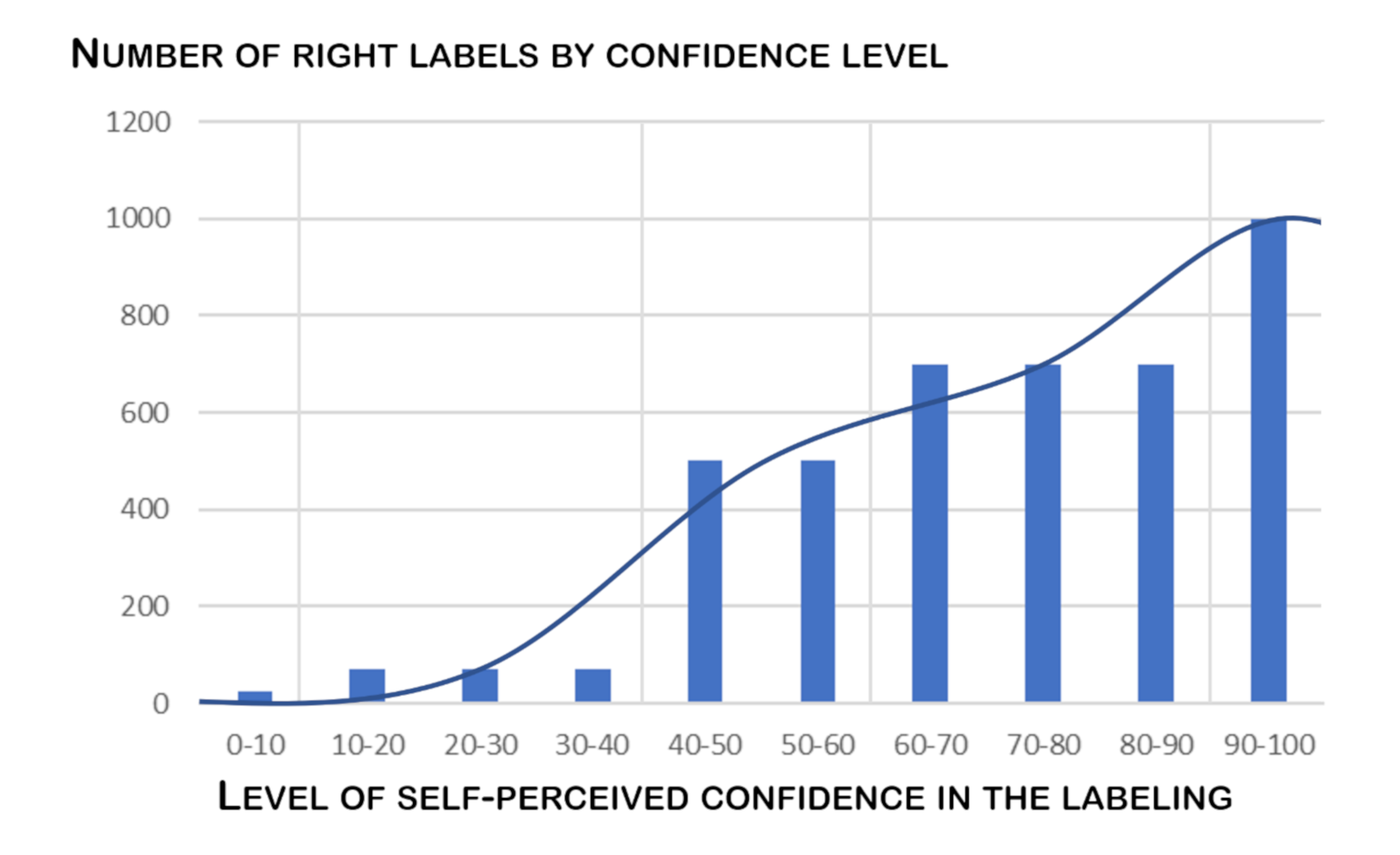}
    \caption{The distribution of the reported confidence levels with respect to the number of cases correctly classified by the 13 radiologists.}
    \label{fig:districonf}
\end{figure*}

In Figures~\ref{fig:res},~\ref{fig:ROC},~\ref{fig:scatter} and \ref{fig:districonf} we graphically report the main elements collected from the two user studies described in Section~\ref{sec:methods}. In particular, Figure~\ref{fig:res} report the minimum acceptable diagnostic accuracy that the involved clinicians deemed necessary to have a useful support; and the preferences about what kind of accuracy they deem more important at class level

In Figure \ref{fig:districonf} is shown the distribution of the reported confidence levels for the cases that had been classified correctly. Notice that, as expected, the majority of the correctly classified cases was associated with a medium to high confidence level. 

In Figure~\ref{fig:scatter} is shown the complexity of each case (a single circle in the figure), related to the average confidence of the raters and their accuracy for that specific case. 

Figure \ref{fig:ROC} shows the radiologists' performance at the labelling task (with respect to the MRNet Gold Standard) and the ROC curve of a ML model, namely a Convolutional Neural Network, that we developed on the \emph{MobileNet} architecture \cite{howard2017mobilenet}, originally trained on the ImageNet dataset and then re-trained via transfer learning on the full MRNet training set (except the 417 instances considered in this study, which were then used to evaluate the performance of our ML model). This model achieves an Area under the ROC curve (AUROC) of .89 and an accuracy of 0.84 (see \textit{a} in Table~\ref{tab:accuracies}). The \textit{mrnet-baseline} model developed by the Stanford University (still the best to date on this dataset) achieves an AUROC of .91, and an accuracy of 0.85~\cite{bien2018deep}.

The accuracy of the performance of each rater is also reported in Table~\ref{tab:acc} more analytically. Moreover, from the first study we got that the preferred average minimum accuracy for a diagnostic support was 93.8 [CI: 92.4,95.3]: incidentally, we also asked the respondents about the minimum \emph{prognostic} accuracy (that is the capability to predict the patients' health status in the future) and this resulted to be almost ten points lower; this probably reflects the fact that human raters are much worse in predicting the future than interpreting present conditions~\cite{white2016systematic}.

\begin{figure*}[tbh]
    \centering
    \includegraphics[width=\textwidth]{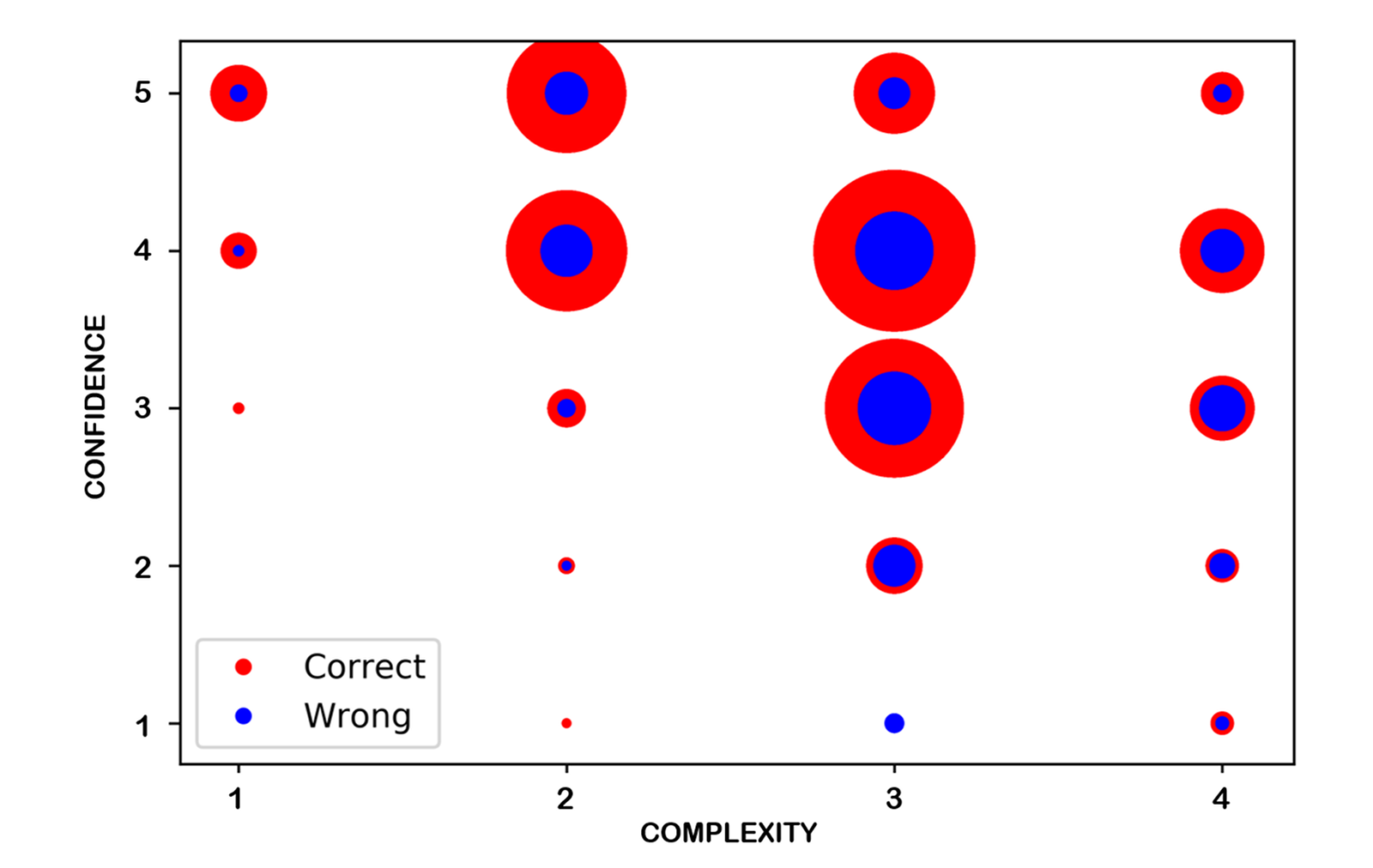}
    \caption{A discrete scatter plot showing the proportion of right (red) and wrong labels (blue) as a function of the discrete levels of confidence and complexity reported by the raters.}
    \label{fig:scatter2}
\end{figure*}

This latter result can inform the setting of $\tau$ and hence the evaluation of \emph{confident accuracy}. Thus, with reference to Figure~\ref{fig:districonf}, we report four \emph{confident accuracy} scores on the basis of different values of $\tau$: $\tau = 0.6$, that is the minimum confidence level reported by radiologists which corresponds to the maximum number of correct answers; $\tau=.75$, that is the confidence associated with the majority (cumulated) of right answers (i.e., the 50.1\% of the area under the integral in Figure~\ref{fig:districonf}, from the right); $\tau=.8$, that is the the confidence level reported by the radiologists corresponding to the quarter of correct answers with highest reported confidence (that is the answers for which the risk of guessing game is the lowest); and $\tau = 1$, that is the maximum level of confidence.

Another interesting result from the first user study regards the preferences about the relative importance to properly detect a class with respect to the other(s): more than two thirds of the respondents claimed that they would find more useful a diagnostic support that could be optimized to be either \emph{more specific} (37\%), and hence better at avoiding false positive errors, or \emph{more sensitive} (37\%) (and hence better at avoiding false negative errors), rather than having a balanced support (11\%). This means that more than two thirds of the sample of respondents expressed a preference for models that weight errors differently according to their type, and that are better at avoiding a kind of error more frequently than the other.

This gives motivation to consider priority in $Ha$ and suggests that \emph{prioritized accuracy} could differ from regular accuracy according to values of $p$ different from $0.5$, that is in all of those cases where instances should be weighted according to the importance attached to avoiding false positives or false negatives, and $p$ should be therefore set at $\langle 0.75, 0.25 \rangle$ or $\langle 0.25, 0.75 \rangle$, respectively. On the other hand, the second user study, the radiological one, allows to set $p$ indirectly, deriving it from the performance points in the ROC space (see Figure \ref{fig:ROC}). In analytical terms, Formula~\ref{eq:priority} can be applied to obtain $\mathbf{p}(0) = 0.52, \mathbf{p}(1) = 0.48$.

\begin{figure}[tbh]
    \centering
    \includegraphics[width=\textwidth]{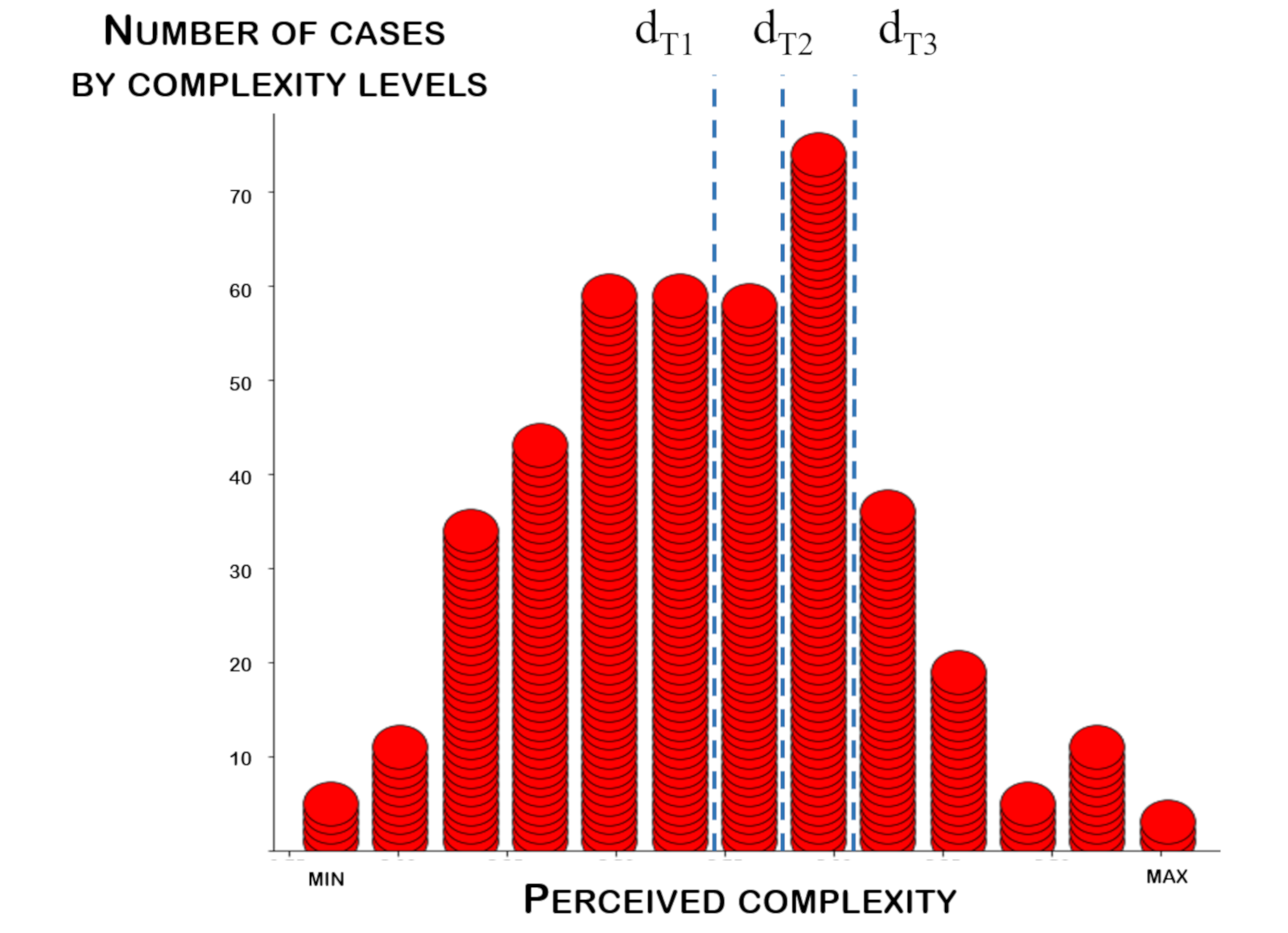}
    \caption{The number of cases grouped by complexity level: $d_{T1}$, $d_{T2}$ and $d_{T3}$ represent complexity thresholds above which lies the 50\%, 33\% and 20\% of the cases, respectively.}
    \label{fig:integral-complex}
\end{figure}

From the radiological study we can also extract a new characteristic of the MRNet dataset: the complexity of its cases. This allows us to compute the \emph{practical accuracy} of our ML model (i.e. $Ha(\mathbf{d}) := Ha(m | \frac{1}{|C|}, \frac{1}{|C|}, \mathbf{d})$). We also considered the practical accuracy obtained when applying a binary threshold $d_t = 2.75$ to the reported complexity values, i.e. setting $\mathbf{d}_0(x) = 1$ for all cases with average reported complexity greater than $2.75$ and $0$ otherwise. The particular threshold was chosen to obtain approximately an equal number of difficult and non-difficult cases (see $d_{T1}$ in Figure~\ref{fig:integral-complex}). In the Section~\ref{sec:discussion}, we will discuss other thresholds.

\begin{figure}[bth]
    \centering
    \includegraphics[width=\textwidth]{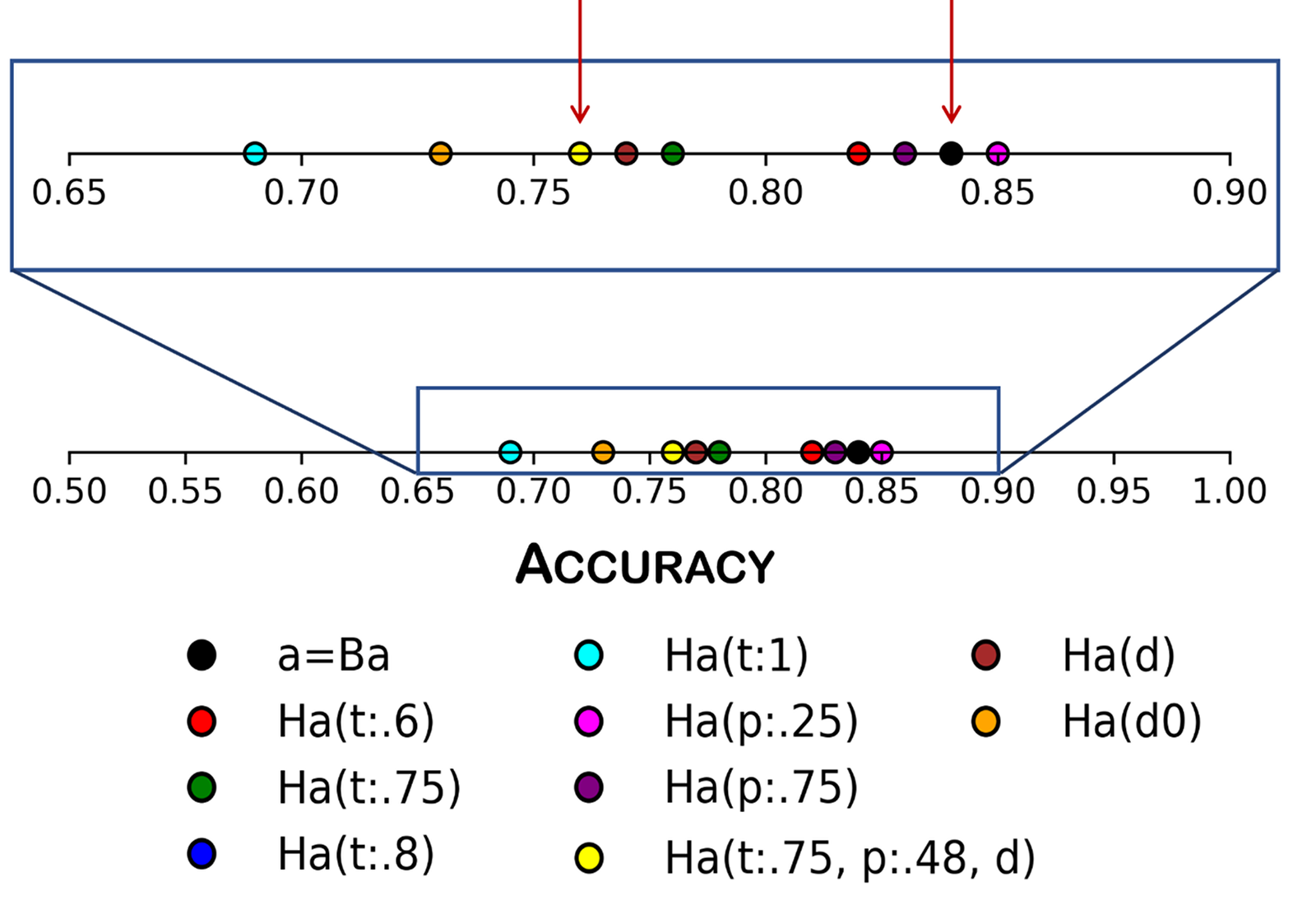}
    \caption{The different accuracies (regular, balanced and different measures derived from the $Ha$ metrics) achieved by our predictive model with respect to abnormal/normal classes on a sample of the MRNet dataset. Regular accuracy and a composite estimate of the H-accuracy are highlighted.}
    \label{fig:accuracies}
\end{figure}

\begin{figure*}[!ht]
\centering
  \includegraphics[width=\textwidth]{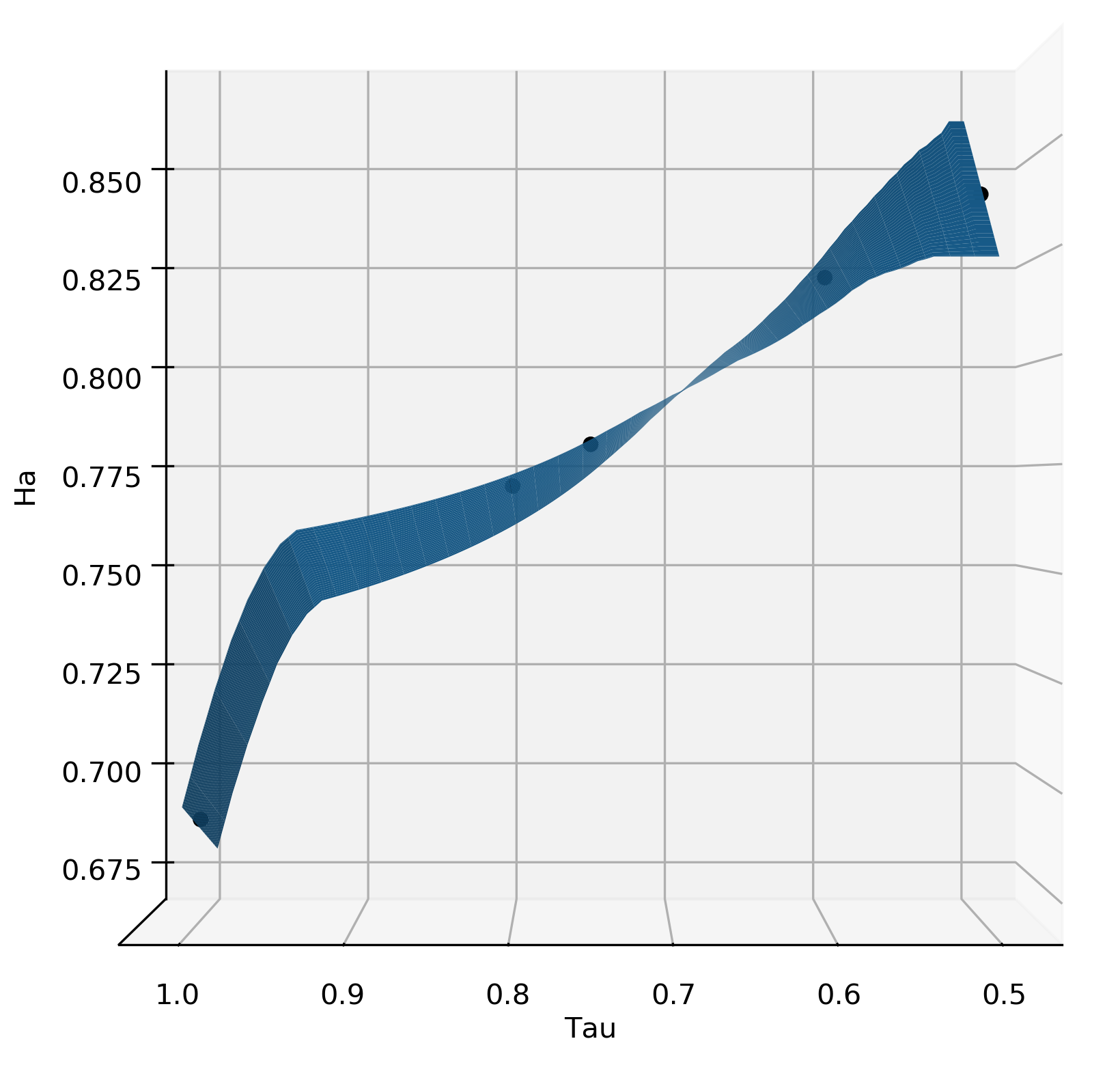}
  \caption{The values for the \emph{confident accuracy} at different confidences $\tau = .5, .6, .75, .8, 1$.}
  \label{fig:tau}
\end{figure*}

\begin{figure*}[!ht]
  \includegraphics[width=\linewidth]{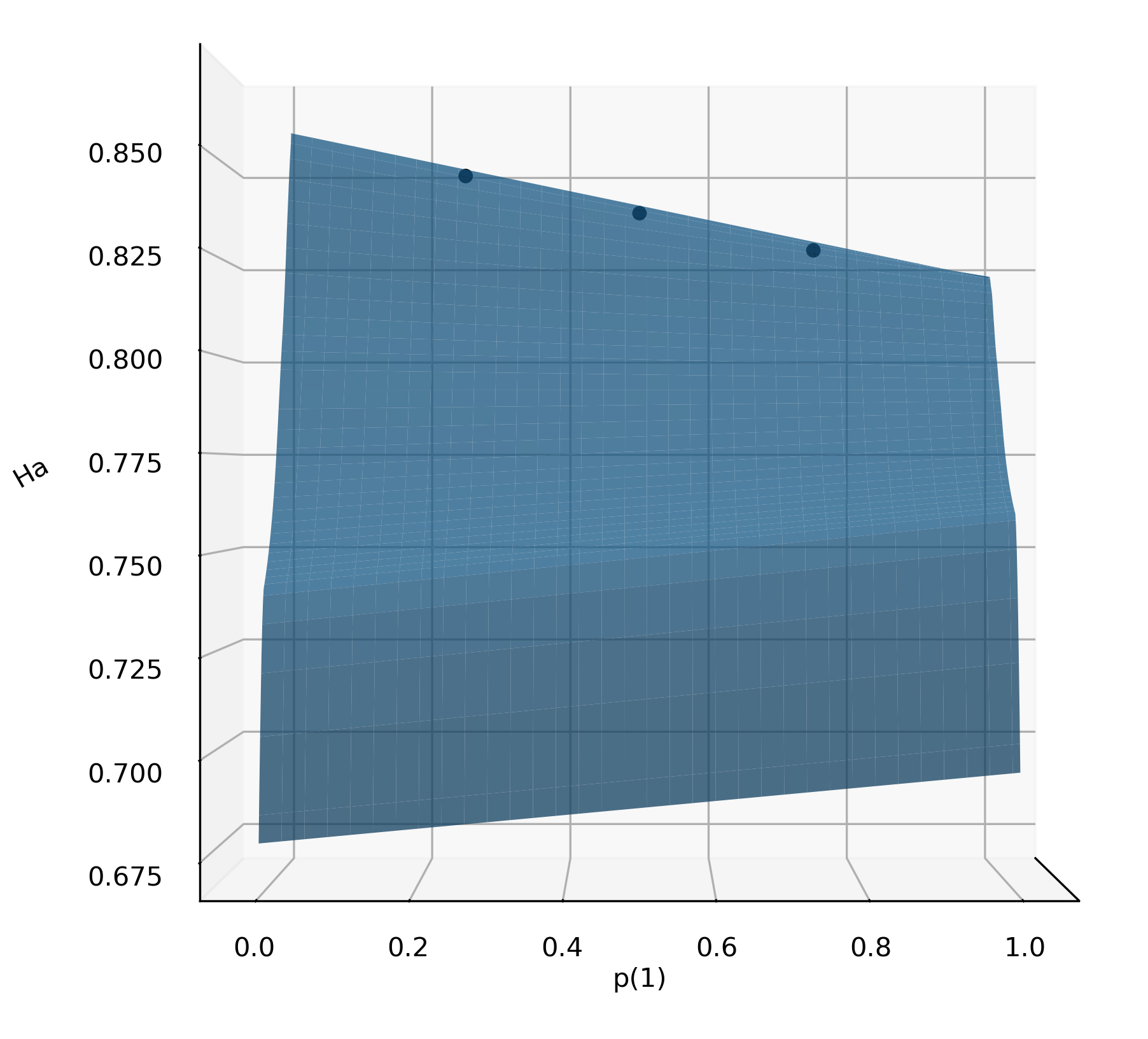}
  \caption{The values for the \emph{prioritized accuracy} at different values of the priority for the positive class $\mathbf{p}(1) = 0.25, 0.5, 0.75$.}
  \label{fig:priorities}
\end{figure*}

Table \ref{tab:accuracies} reports all the accuracy scores mentioned above and derived by Formula~\ref{eq:ha}, all together with the \emph{balanced accuracy} and the \emph{regular accuracy}, as obtained by evaluating the described ML model on the same 417 cases which were also given to classify to the 13 radiologists; Figure~\ref{fig:accuracies} shows the confident and prioritized accuracies (along with balanced accuracy, $Ba = Ha(\tau = 0.5) = Ha(\mathbf{p}(1) = 0.5)$) achieved by the ML model by visualizing them on the $\tau$ and $\mathbf{p}(1)$ projections of the whole 3D visualizations of the $Ha$ surface profile. Comparing Figure~\ref{fig:accuracies} with the values reported in Table~\ref{tab:acc}, it can be seen that the accuracy of the model is only slightly higher than the average accuracy across the 13 radiologists. We can, furthermore, observe that as argued in Section \ref{sec:introduction} the regular accuracy $a$ tends to overestimate the real performances of the model: indeed, in all cases (except the balanced accuracy $Ba$, whose equality with $a$ can easily be explained by the fact that the two classes were near perfectly balanced; and the $Ha(\mathbf{p}(1) = 0.25$, which puts greater importance on the slightly majority classes $l = 0$) the instances of the $Ha$ metrics had values strictly lower than the regular accuracy $a$. In fact, we can see that the \emph{confident accuracy} $Ha(\tau)$ monotonically decreases with increasing $\tau$ (this also holds for the $Ha$ measure in general, as can be seen in Figure \ref{fig:tau}) and the same observation can be made with respect to the practical accuracy $Ha(\mathbf{d})$ which easily shows that the ML model was evidently better at classifying the simpler cases than the more complex ones. Also, we can observe from Figure \ref{fig:priorities} that on our sample dataset the $Ha$ increases when a greater priority is assigned to the negative class $l = 0$ (no abnormality), which can be easily explained by the observation that the dataset presented a small skew in favor of this class.

\begin{table*}[!ht]
    \centering
    \setlength\tabcolsep{2.5pt}
    \begin{tabular}{|c|c|c|c|}
    \hline
         
         \multicolumn{2}{|c|}{a = Ba = $Ha(\mathbf{p} : .5)$} & $Ha(\mathbf{p}: .25)$ & $Ha(\mathbf{p}: .75)$\\\hline
         \multicolumn{2}{|c|}{0.84} & 0.85 & 0.83\\\hline\hline
         
         $Ha(\tau: .6)$ & $Ha(\tau: .75)$  & $Ha(\tau: .8)$ & $Ha(\tau: 1)$\\\hline
         0.82 & 0.78 & 0.77 & 0.69\\\hline\hline
         
         \multicolumn{2}{|c|}{$Ha(\tau: 0.75, \mathbf{p}: 0.48, \mathbf{d})$} & $Ha(\mathbf{d})$ & $Ha(\mathbf{d}_0)$\\\hline
         \multicolumn{2}{|c|}{0.76} & 0.77 & 0.73\\\hline

    \end{tabular}
    \caption{Accuracies (regular, balanced and different measures from the $Ha$ metrics) achieved by the ML model with respect to abnormal/normal classes on a sample of the MRNet dataset (containing only normal cases and ACL or meniscal tears).}
    \label{tab:accuracies}
\end{table*}

\section{DISCUSSION}
\label{sec:discussion}

In Section~\ref{sec:methods}, we provided proof that \emph{H-accuracy} is a robust measure of accuracy and demonstrated it is suitable for the assessment of the performance of machine learning models. This metrics requires to set two parameteres $\tau$ and $p$ and determine $d(x)$ that is the complexity of the cases in the dataset used to assess the model.

In this section, we will share some reflections on the different kinds of accuracy scores that Formula~\ref{eq:ha} allows to produce, and on the methods that can be applied to set the above parameters, to get results like those reported in Table~\ref{tab:accuracies} and Figure~\ref{fig:accuracies}. 

\subsection{Confidence, for reliability}

First, let us consider \emph{confident accuracy}. This is defined according to values of $\tau$, and is the accuracy of a system that is at least \emph{$\tau$ confident} of its output, such that we can claim that ``it does not take guesses''. Thus, $\tau$ is the threshold of the confidence (in the membership of instance $x$ in any class $l$), below which, even if it classifies $x$ correctly, it is still penalized.

\emph{Confident accuracy} finds motivation in how human experts behave: in situations of uncertainty they ``don't usually toss a coin'' ($\tau$ = 50\%) but rather abstain, at least temporarily, and either look for stronger clues or evidence~\cite{lin2015addressing}, or refer to the second opinion of a more expert colleague~\cite{palmieri2017second}. When multiple options are available (i.e., in multi-class tasks), experts can choose the option (e.g., the diagnosis) that is more likely or plausible among the ones available, but they would still act as the option were true only when the plausibility is high enough (otherwise, at least in medicine, a ``wait and see'' attitude is generally recommended~\cite{grady2010less}). 

Although there are some machine learning applications that apply abstention in conditions of uncertainty (e.g~\cite{campagner2019three,campagner2019exploring}), this approach is not commonly adopted, and hence it is common that decision support systems have users consider recommendations that are associated with low prediction scores and have a higher chance to be wrong, often without their knowledge, and hence mislead them. 

Setting a $\tau$, and reporting a corresponding confident accuracy ($Ha(m|\tau)$, can help users evaluate this risk; for this reason we recommend to compute (and communicate) several points of \emph{confident accuracy}, like in Figure~\ref{fig:tau}  to understand the magnitude of the divergence between regular accuracy and \emph{confident accuracy}.

On a qualitative basis we recommend setting this acceptability threshold at 51\% for multi-class problems, and higher for binary tasks. In this latter case, $\tau$ can be set according to different criteria. 
If confidence is recorded, $\tau$ can be the lowest confidence level expressed by the involved raters that is associated with a sufficient cumulative amount of right answers (e.g., at least 80\%): in our case the corresponding $\tau$ was 0.6; alternatively $\tau$ can be the average confidence of the raters, or the average confidence of the best (i.e., most accurate) rater expressed during the process, depending on whether the prospective decision makers need a highly confident (and hence reliable, in some way) advice from their decision support or not. 

Setting a high $\tau$ and relying on $Ha$ instead of regular accuracy raises the bar for the performance of machine learning in real-world settings and reflects the requirement to have really accurate models that minimize ``guessing'' behaviors in a sensitive context. 

\subsubsection{Two sides of a threshold}

In light of the relationship between \emph{H-accuracy} and the \emph{Net Benefit} demonstrated in Section~\ref{sec:methods}, we can see that a significant difference between the two measures 
lies in a different semantics attached to the probability threshold $\tau$. The semantics of $\tau$ in $Ha$ is specified by the penalty function, $\sigma$; therefore adopting a specific formulation for this sort of weighting function of right predictions allows to assimilate the two semantics. 

Let's see if the equivalence found by Theorem~\ref{th:nbha} makes sense for the practical evaluation of decision support models. 
As hinted above, in the formulation of $NB$ (as well as in ROC analysis), the $\tau$ is intended as a non-symmetric \emph{risk} threshold \cite{zhang2018decisioncurve} associated with the risk of the case at hand being a true positive and hence requiring a treatment. In the $NB$ framework, the above asymmetry lies in the higher importance of the condition ``being a true positive'' (and the related costs of missing it): in this light, $\tau$ can also be seen as the probability threshold above which the model \emph{must} classify instance $x$ as positive (1) to bring benefit, and below which it must state the opposite (obviously). Thus, while in the formulation of $Ha$ given in Equation \ref{eq:ha} $\tau$ is set equal for all classes, for $NB$ this is set for the positive class, and $1-\tau$ holds for the alternative option. 

Also $\tau$ in the $NB$ can be set according to a number of criteria (except the optimization of net benefit itself~\cite{kerr2016assessing}): most of the time this choice is driven by an intuitive assessment of the relative harm of false positives and false negatives with respect to the magnitude of both costs and benefits (i.e., $\tau=costs/(benefits + costs)$\footnote{In this footnote we give informal proof of this equivalence in the scenario of establishing whether a medical intervention (e.g., a treatment) should be performed or not. Let us then denote with $a, b, c, d$, respectively, the utilities related to the possible events \emph{disease + treatment}, \emph{no disease + treatment}, \emph{disease + no treatment}, \emph{no disease + no treatment}. Then, the $NB$ can be derived~\cite{peirce1884netbenefit,vickers2006decisioncurve}, by assuming that $\tau$ is a probability threshold at which a patient would be uncertain between accepting the treatment or not (that is between being considered at high-risk or not~\cite{kerr2016assessing}), by setting $a - c = 1$ (i.e. the utility of a true positive) and solving for $b - d$ (i.e. the utility of a false positive) in $\frac{a - c}{d - b} = \frac{B}{C} \frac{1 - \tau}{\tau}$, where $B$ (resp. $C$) is the benefit (resp. cost) of undergoing treatment, thus $\tau = \frac{C}{B + C}$.
}), which often depends on individual preferences and contextual factors, including the prevalence (that is occurrence in the \emph{observed} population) of the positive class (i.e., $\tau=\pi$). 

On the other hand, the formulation of $Ha$ given in Equation \ref{eq:ha} adopts $\tau$ as a symmetric \emph{confidence} threshold, that is as a sort of \emph{credibility score} that penalizes low confidence in both classes equally, thus rewarding models that minimize the number of predictions which are correct only by chance. This behavior is particularly meaningful when the two classes have similar consequences and outcomes, for instance when both classes correspond to different conditions that have similar costs of false positives and negatives, and when we are primarily interested in obtaining classifiers whose all predictions are highly confident, as a proxy for reliability. 

As anticipated above, $Ha$ can be made similar to $NB$ unless for a multiplication factor, through a specific $\sigma$ function (e.g. see Equation~\ref{eq:sigma_alt} for a function that mirrors the risk-based behavior of $NB$).

\begin{figure*}[!ht]
    \centering
    \includegraphics[width=\textwidth]{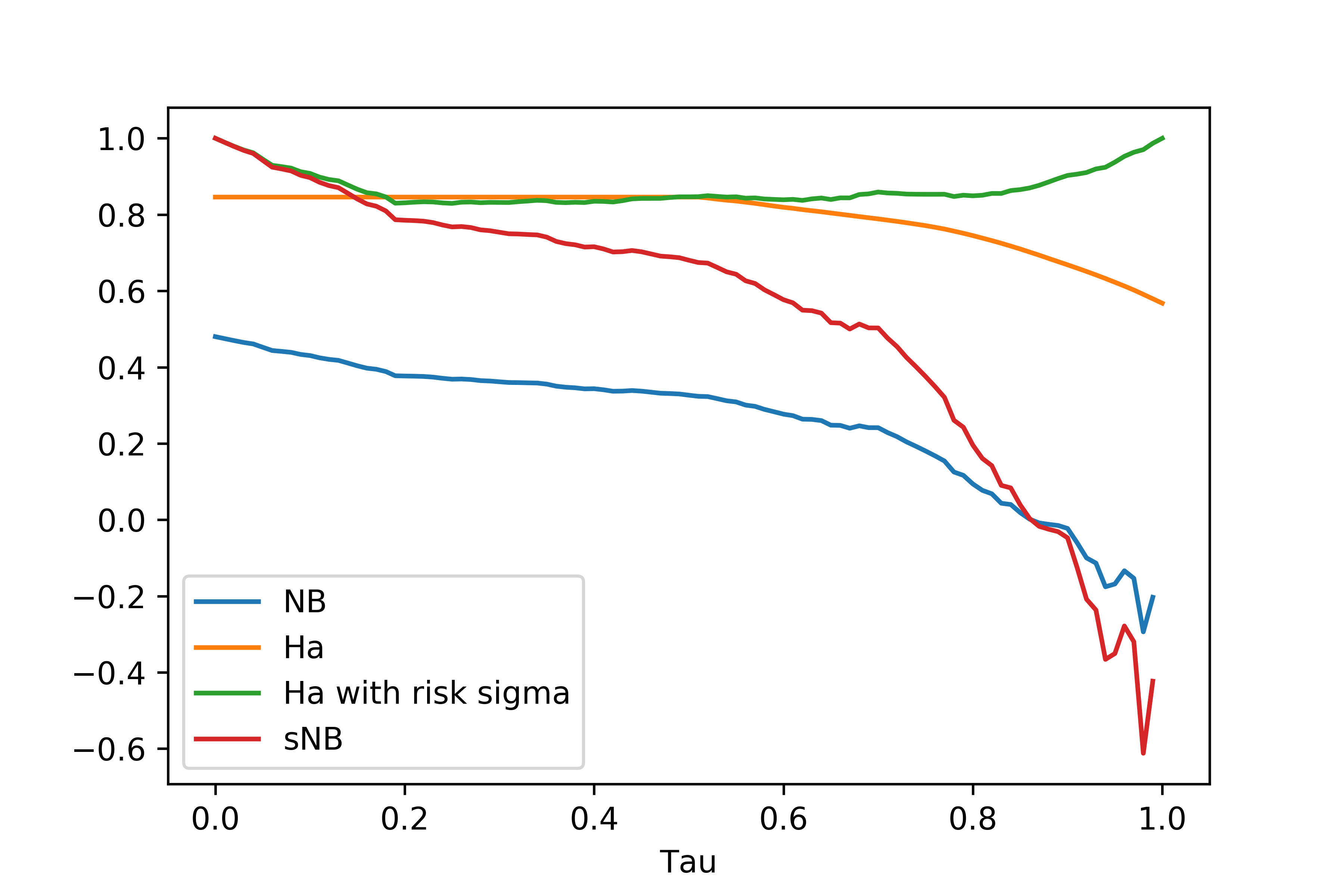}
    \caption{The behaviour of $NB, sNB, Ha(m|\tau), Ha(m|\tau,\sigma_{risk})$ with respect to $\tau$ is illustrated. In the graph of the $Ha(m|\tau,\sigma_{risk})$ we can observe that the score goes to 1 when $\tau$ goes to 0 or 1. This behaviour depends on the fact that using the \emph{priorities} specified by Theorem \ref{th:nbha} changing $\tau$ reduces the importance of one of the two classes (namely, when $\tau = 0$ cases in the negative class are completely ignored, when $\tau = 1$ the same holds for the positive class).}
    \label{fig:nb-vs-ha}
\end{figure*}

An illustration of this relation is shown in Figure~\ref{fig:nb-vs-ha}, which reports the behaviour of $NB$, $sNB$, $Ha(m|\tau)$ and $Ha(m|\tau,\sigma_{risk})$ (see Equation \ref{eq:sigma_alt}) to varying of the value of $\tau$. As regards $Ha$, since $\tau$ acts as a symmetric confidence threshold, the value of the accuracy monotonically decreases when $\tau$ increases because we increasingly penalize non-confident predictions. On the other hand, for $NB$, $sNB$ and $Ha(m|\tau,\sigma_{risk})$ $\tau$ acts as a risk threshold. When it is low (resp. high) there will be a preference for treatment (resp. no treatment) and a lower (higher) weight will be attached to the true positives (resp. true negatives), thus resulting in a symmetric curve. In the $NB$ and $sNB$ measures, the risk threshold reflects the cost associated with false positives, which exponentially increases to increasing of $\tau$, thus driving towards negative scores: in this sense, these two measures are appropriately interpreted as the \emph{utility-based} indicator of the value of a predictive model.

On the other hand, $Ha$ can be considered a \emph{usefulness-oriented} indicator of the value of a predictive model, as well as an indicator of accuracy, as the name suggests. The two concepts are obviously related, but different and, we claim, tightly \emph{complementary}: expressing in other words the result of Theorem \ref{th:nbha}, we can say that H-accuracy, as any accuracy estimation, yields a probabilistic estimate of the discriminative power of a model; when this is discounted by the uncertainty acknowledged at the core of the discriminative process (1-$\tau$, that is but an estimate of what separates us from making certain predictions), we also get an estimation of the net benefit this model can yield if adopted in human decision making.

\subsection{Priority, for tailorability}

As we said in the previous section, a cost-benefit analysis assumes that the posituve class is more important than the negative one. In the $Ha$ formulation, class priority can be fully customized. By setting different class priority, $p$, \emph{prioritized accuracy} can be defined. Values of $p$ can be seen as a sort of weight attached to rightly predicting one class over the others, and can be set naively: for instance, if $Ha(m|p)$ must express a clear preference for specificity, $p$ could be associated with $0.75$ (and $0.25$ with respect to sensitivity); and vice versa in case of a preference for sensitivity. Otherwise, if no preference is expressed, a balanced weighting ($0.5$~--~$0.5$) can be adopted. 

On the other hand, priority $p$ can be set empirically in two ways: either explicitly or implicitly. The explicit way entails to get the prevalence of each class in the reference population or to collect the preferences of a panel of experts for the classification task at hand by means of a psychometric questionnaire. In this latter case, radiologists could express different class priorities for different exams (x-rays, MRI, CTs), for different pathologies (fractures, lesions, tumors), and differently from other specialists, like orthopedists and oncologists. In the former case, class priority can be set according to class prevalence $\pi$, or its complement ($1-\pi$) if rare cases should be considered less (more) important to detect, respectively. On the other hand, the implicit case occurs when experts are involved as dataset labelers, and $p$ is set according to their performance by means of Formula~\ref{eq:priority}. 

A last way to define $p$, in the binary case, is suggested by the relation established between $Ha$ and the $NB$. Indeed, in light of the utility-based definition of $\tau$ given in Section \ref{sec:methods}, $p$ can be defined as a function of both a particular value of $\tau$ and of the prevalence of the positive class $\pi$ as $p = \alpha*\langle \tau*(1 - \pi), (1 - \tau)*\pi \rangle$ where $\alpha$ is a normalization factor.

\subsection{Complexity, for usefulness}

As said in the previous section, class priority is a way to embed class importance into an accuracy metrics. On the other hand, the function $d(x)$ is a way to consider particular types (implicitly so) of instances important. For this contribution we chose to limit ourselves in considering important complex cases, that is the cases for which a decision support could be more useful. In particular, with the term \emph{complexity} we here refer to the decision effort and difficulty to classify an instance of the test set. We can have a visual rendering of complexity, naive as it may seem, by looking at Figure~\ref{fig:ROC}: if classification experts are involved (like radiology diagnosticians), it is the difficulty of the cases that they reviewed that prevents the points representing their performance to be all close together and piled up at the top left corner of the ROC space.

However, by looking at the diagram and looking up the accuracy data reported in Table~\ref{tab:acc}, the reader should resist the temptation to compare the radiologists involved in the MRNet labeling task the machine performance at the same task: in this paper, we are making the case of providing multiple accuracy measures to describe the performance of a classification model in a more informative manner; we are \emph{not} suggesting that a specific model outperformed human decision makers, as claimed in many paper recently~\cite{Liu2019,shen2019artificial}. This is not the case because the radiologists involved in this study reviewed the imaging that was made available in the MRNet dataset, and not the original, high-res images from which that dataset has been produced, and they did it on standard monitor sets, not on calibrated screens. Moreover, the radiologists were obviously aware that they were involved in an experimental setting: although we took care in explaining that the task required their maximum attention not to put noise into the dataset labelling, also a \emph{laboratory effect} should be factored in~\cite{gur2008laboratory}, that is the fact that their performance in this experiment should not be considered representative of their reading skills in real-life settings at any extent. 

But how should $d(x)$ be determined, and hence the \emph{practical accuracy} score? Here, we recall that $d(x)$ is a function returning a complexity score for each case $x$, that is how hard it is to correctly classify the case. This function can be determined in different ways, among which we can distinguish between direct and indirect methods. In regard to the former approach: we can ask the doctors involved to express whether they found a case $x$ (from the test set, or the whole dataset under consideration, either at ground truthing or also subsequently, on an already established gold standard) difficult or not (0 or 1), as if we asked them if that case deserves to be shared with a class of residents or trainees, or even with the specialist community, as a case report~\cite{ortega2017importance}. Similarly, we could ask them to rate how difficult they perceived the case $x$ on an ordinal scale. This kind of rating can also be used to distinguish the cases between simple ones (still, 0) and complex (1), on the basis of a given threshold $d_{T}$, beyond which a case is considered to be complex and difficult to interpret. 

The threshold $d_{T}$ can be fixed a priori, like $d_{T}=0.5$ on a normalized scale, or on the basis of the ratings collected empirically. For instance, with reference to Figure~\ref{fig:integral-complex}, the threshold of complexity can be set at the average case complexity ($d_{T}=0.275$), or in such a way that only the most difficult portion of cases are considered. What portion of cases to consider is an open issue, though. In our case study, we considered the 50\% of cases, not to reduce the test set too much. However, in another study (unpublished), we extracted 1153 wrist x-rays from the more than 47000 wrist radiographs taken at the Galeazzi institute in the 2002-2018 period: in this case, the cases considered complex (with comminuted fractures) by a top-performance radiologist were exactly one third (33\%), while the simple ones (with compound fractures) were the 20\% of the sample (with a 47\% of negative images). On the other hand, others~\cite{berner2014can} suggest to be even more selective: ``the 80~--~20 rule has been said to apply to the cases an experienced physician sees in his or her practice. That is, about 80\% of the patients have conditions that are fairly easily diagnosed. For this large majority of cases, not only are clinical decision support systems not needed, using them would be an inefficient use of the physician’s time and might even lead to expensive and unnecessary additional diagnostic testing.''. Thus, by choosing $d_{T}$ to be 0.5, instead of 0.33 or even 0.2 in our user study, we chose a conservative estimation for $d(x)$ (see Figure~\ref{fig:integral-complex}).

Alternatively, once a gold standard is available or has been established, also indirect methods are possible; for instance $d(x)$ can be set on the basis of the error rate of the raters involved for that specific case $x$: the higher, the more difficult the case is; or on the basis of the confidence the raters express in assessing the case, irrespective of whether their answer was correct or not (the lower the confidence, the higher the complexity~--~see Figures~\ref{fig:scatter}~and~\ref{fig:scatter2}); or, still, by combining accuracy and subjective confidence together. Additionally, also the average accuracy of the raters (that is their accuracy on the whole dataset), seen as a proxy of proficiency, can be used to modulate the subjective perception of complexity: if a case is perceived difficult by a top performer its complexity should be considered higher than if only the low-performers considered it such. Proficiency, in its turn, can be assessed by self-assessment, or by combining basal (self-)assessment and task performance. Many approaches can be pursued: future research is then necessary to understand what approach makes more sense in different clinical settings.

Above, we hinted at the use of ordinal scales to let raters assess the complexity of a case. The choice of a proper scale is far from being a trivial task. In designing our user study, we considered two alternatives, but adopted one of them eventually: a first approach that we could call \emph{semantic}, and an alternative one, \emph{pragmatic}. The semantic rating scale was conceived as a traditional 6-value semantic differential ranging from 1 (``very easy case'') to 6 (``very difficult case''). This choice sounds reasonable as this scale is common in many psychometric assessments. However, by discussing with the radiologists, we observed that having a group of medical raters agree on a common way to assess case complexity homogeneously is a hard, if not over-ambitious, task: the resulting ratings would be affected by a too large amount of arbitrariness, and hence noise; to put it naively, ``it's easy what you know how to do; the rest is difficult''. 

Therefore, we preferred to adopt the pragmatic scale: to trace back relatively few complexity levels (in our case 4), to the \emph{reasonable expectations} of the raters in regard to easy-to-delimit professional categories and their exam reading proficiency. In particular we identified the following categories: non-specialist doctors, associated with level 1; residents (as specialist-to-be doctors), as level 2; specialists within a broad specialty (like radiology - level 3); and sub-specialists (as specialists who further specialize in a kind of disease or exam, like orthopedic radiology, or even more narrowly, MRI orthopedic radiologist), as level 4. To avoid the reasonable objection that not all specialists are the same, we formulated the ordinal levels by referring to the capability of a ``large majority'' of people from a professional category to correctly classify a specific case. Consequently, also a fifth category could be considered: cases that ``only a few sub-specialists, the very best ones, could solve'' (level 5).

Why assessing complexity is important, and therefore \emph{practical accuracy} is? Also in this case, it is \emph{naturalistic decision making}~\cite{klein2008naturalistic} that motivates this idea. Think of a very accurate decision support for trivial or easy cases: the medical experts would not consult it, or if they do, would neglect and override it with no regret or doubt in case they disagreed with it\footnote{Here let us just hint at the impact of these ``easy misses'' on doctors' trust and on the perceived reliability of the support.}. Conversely, complex cases are those for which even experts would like to receive confirmation or, if lost about them, a proper aid. In these cases, right advice can have a strong impact on the final decision, and wrong advice can mislead the decision makers, a situation that could be worsened by the automation bias mentioned above (i.e., over-reliance on the support, which could be induced by a higher, almost oracular, accuracy exhibited by the system on the easy cases). 

For these reasons, we deem the \emph{practical accuracy} an important measure to compute and report, to understand the real usefulness of a decision support in naturalistic (i.e., real-world) decision making by expert practitioners. Consequently, further research should be aimed at understanding what the best way to represent medical complexity (i.e., $d(x)$ in $Ha$) is, and also whether this construct should be integrated with some related information, like \emph{relevance}. Intuitively, this latter dimension could relate to similar but slightly different concepts, like rarity, significance, seriousness, or the impact of missing the right interpretation for the patient's recovery\footnote{These are nuances that the H-accuracy allows to specify more accurately than what the Net benefit construct allows}. An evaluation, expressed by experts, of these further concepts could be transparently integrated in the definition of $d(x)$, thus obtaining a multi-faceted representation of the importance of correctly classifying specific cases by simply combining (e.g. via linear combination) all these different factors to obtain a single, weighting function mirroring the needs of the clinical practitioners.

Two limitations of this study regard complexity: first, the involved radiologists did not evaluate the complexity of the whole MRNet dataset, but of a sample of 417 cases, randomly drawn from the former set. Second, they did not assess the complexity of the original imaging from which the MRNet dataset was derived. 

\begin{figure*}[!ht]
    \centering
    \includegraphics[width=\textwidth]{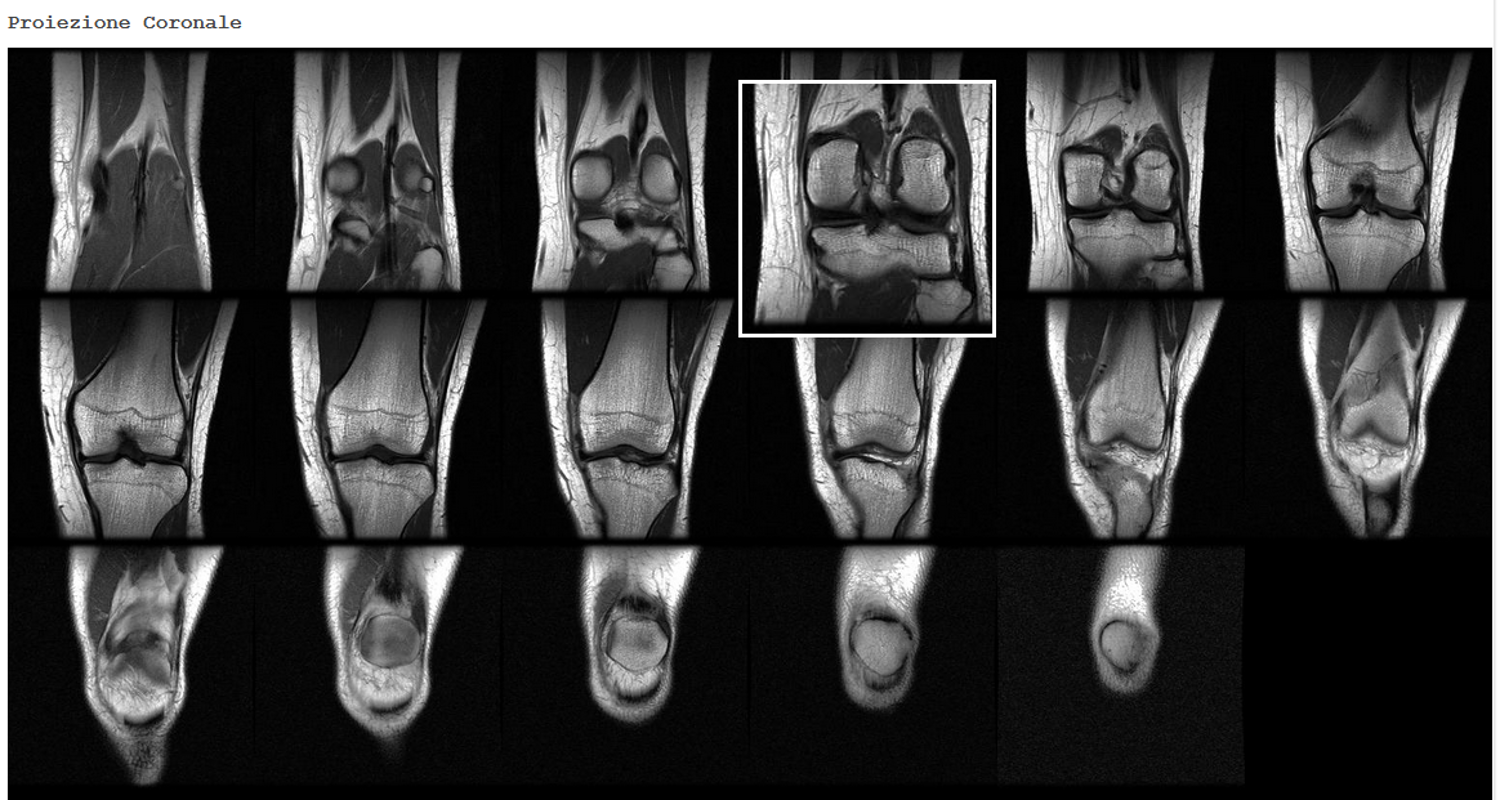}
    \caption{An example of how the radiologists could review the MRnet imaging in the online questionnaire platform we developed for this study.}
    \label{fig:image}
\end{figure*}

In particular the doctors reviewed all the imaging in three projections, (i.e., axial, coronal, sagittal planes); each plane was associated with a single picture 1920-pixel high and 1152-pixel wide that was composed of a varying number of smaller pictures, the single ``cuts'', up to 48. The platform allowed the doctors to inspect each image with a ``magnifying glass'', a squared area of 200 x 200 pixel, like in Figure~\ref{fig:image}.

In regard to the latter point, we conjecture that a low-quality imaging would make reported complexity higher in absolute terms than in a high-quality setting, but likely not too different in relative terms. Since the equation of H-accuracy (see Formula~\ref{eq:ha}) normalizes complexity scores, we deem this limitation having low impact on our argumentation. Likewise in regard to the former point, the random selection of the cases from the MRNet Knee MRI Dataset (417 out of 1370) should keep the margin of error below the conventional threshold of 5\%, and make our argumentation affected by this limitation only marginally, if any.

As a bottom-line reflection, let us look at the different accuracy scores reported in Table~\ref{tab:accuracies} and Figure~\ref{fig:accuracies}. The best model on the MRNet dataset (a deep learning neural network~\cite{bien2018deep}) exhibits a regular accuracy of 0.85. Our deep learning model, tested on a subset of that dataset, is only slightly worse: 0.84. Since our test set was balanced with respect to abnormal/normal cases, the regular and balanced accuracy of our model coincide. 

The performance of both predictive models seem much lower than what the doctors involved in our first user study expressed as the minimum accuracy to have a \emph{useful} support. Obviously, this depends on the people involved, their expertise and their specialty: we conjecture that oncologists could express different preferences than orthopedists, and likely each setting could host users with different expectations and attitude towards an Artificial Intelligence support. 

However, the main point of this contribution is another one: the performance of both models are even lower than what depicted by the regular accuracy figure. If we weight \emph{less} the cases in which the models could have just guessed the right answer ($\tau>.75$), we get a worse performance, by even 15 percentage points (if we demand the highest confidence by the model). If we consider the complexity of cases, we get a worse performance by 7 and even 11 percentage points if, respectively, we consider an ordinal (4-value) spectrum of complexity (with our pragmatic scale), or a more common dichotomous distinction.

Different research groups and companies across the world in the domain of machine-learning based decision support struggle to surpass competitors even by a few percentage points in accuracy (or related measures), with a common ``psychological'' threshold to interest the medical stakeholders set to 80\% at least~\cite{floares2009using}: besides these figures, in this paper we made the point that is important to inform stakeholders and practitioners by also showing them alternative accuracy measures, with a \emph{clear clinical and practical meaning}, as H-accuracy aims to be.

\section{Conclusions}
\label{sec:conclusion}

With the increasing need for prospective studies in the field, and third-party certifications of AI-based decision supports (seen by the FDA as \textit{Software as a Medical Device} or as \textit{Medical Device Software} by the European Medical Devices Regulation), competent authorities, regulators, AI manufacturers, distributors, and health-care providers, all need meaningful ways to assess the performance of these systems in clinical practice.  

We are not the first ones to point out how the current use of the accuracy indicator for understanding and reporting how well a classification model works (even assuming a proper testing process producing such an indicator) falls short of communicating the most important feature of a decision support, that is \emph{being useful} in daily practice. 

Indeed, other contributions have discussed the limits of common performance measures, like the AUC~\cite{halligan2015disadvantages,lobo2008auc,pencina2015evaluating} and accuracy itself~\cite{sokolova2006beyond,albacete2014accuracy}) and proposed how to minimize the odds of misinformation about ML model performance, usually by recommending to provide a number of different measures. 

Conversely, we made the point that the medical community (like many others that increasingly rely on predictive models and decision support systems) must have few, summative but yet comprehensive and informative ways to learn about, compare, report on, and communicate about the ``skills'' of Artificial Intelligence in medicine~--~ideally a \emph{single} measure~--~and stick with this, also to allow for fair comparisons. Recently, Mandrola and Morgan in~\cite{mandrola2019important}, to warn of the risks of incidental findings and the related overuse that a highly sensitive decision support could exacerbate, have observed how ``the problem with decision support is that it must be designed \emph{to add value} and be easily accessible without increasing burden for clinicians [\ldots that is it] needs to \emph{better provide relevant information} at the point of care to make decision-making easier for clinicians.'' (our emphasis).
Some author has then recently proposed this one best measure is the \emph{Matthew correlation coefficient}~\cite{chicco2017ten}, which is not affected by class imbalance and is generalizable to multiclass settings.
However, this metrics is not intuitively related to error rate and, mostly important, does not consider the characteristics of the available data, nor the preferences of the intended model users.

For this reason, in this paper we proposed a novel metrics that takes into account the above elements, to provide an indicator of the reliability and value of the potential advice by a decision support. In particular, by providing an analytical formulation of this metrics, we also pointed out meaningful areas of the resulting function to focus on specific aspects of the model performance, like reliability (cf. $\tau$), practicality (cf. $d$) and priority (cf. $p$), and suggested some empirical values to report that we believe could inform the users adopting an ML model exhibiting such skills, namely \emph{confident, prioritized}, and \emph{practical accuracy}. To our knowledge, H-accuracy is the first metrics to go beyond what can be known of a model's performance from the confusion matrix, while still being related to the intuitive notion of ``getting classification right''\footnote{H-accuracy is also ``backward'' compatible, if (in the binary case) $\tau$, $p$ and $d(x)$ are set, respectively to 0.5, 0.5 and any constant value.}.  

However, H-accurcay still regards a \emph{static} assessment of the accuracy of a prediction model, evaluated \emph{holistically} on a whole test dataset, as an indicator of the usefulness of the model. Our future work will explore a more dynamic and case-dependent assessment, to define an indicator of the reliability of the model, not on \emph{any} new case, but on \emph{one particular} new case, specifically: this is the case of a sort of \emph{local accuracy} metrics, evaluated on the basis of the performance of the model on the $n$ (e.g., 10) most similar cases in the test set to the new case to classify.

Once more comprehensive static metrics and convenient dynamic metrics will have been made available to the users, further research should be aimed at evaluating whether and how much the user experience would change and possibly improve when the decision makers can learn these additional information about their decision support~\cite{berner2003diagnostic}, to mitigate automation bias, and increase meaningful use and the overall accuracy of teams of decision makers in real-world settings.   

\section{Acknowledgments}

The authors are deeply grateful to Prof. Giuseppe Banfi, head of the Scientific Department of the IRCCS Orthopaedic Institute Galeazzi of Milan, for the continuous encouragement and the active promotion of the survey reported in the first user study among the employees of the institute. The authors also acknowledge the valuable contribution of Prof. Luca Sconfienza, head of the Operative Unit of Diagnostic and Interventional Radiology at the IRCCS Orthopaedic Institute Galeazzi of Milan, for his help in making the second user study possible, and for involving twelve more valid radiologists in the demanding labelling of the MRNet dataset.

\end{document}